\pdfoutput=1

\documentclass[11pt]{article}
\usepackage[dvipsnames,table]{xcolor}
\usepackage[preprint]{acl}
\usepackage{multirow}
\usepackage{times}
\usepackage{array}
\usepackage[most]{tcolorbox}      
\usepackage{multirow}
\usepackage{latexsym}
\usepackage[T1]{fontenc}         
\usepackage[utf8]{inputenc}
\usepackage{microtype}
\usepackage{inconsolata}
\usepackage{booktabs}
\usepackage{graphicx}
\usepackage{caption}
\usepackage{subcaption}
\usepackage{amssymb}
\usepackage{amsmath}
\usepackage{bm}
\usepackage{makecell}
\usepackage{capt-of}
\usepackage{soul}
\usepackage{textcomp}

\title{RoToR: Towards More Reliable Responses for Order-Invariant Inputs}

\newcommand{\ours}{\textsc{RoToR}}
\newcommand{\sr}{Selective Routing}

\vspace{-0.5cm}
\author{
\parbox{0.8\linewidth}{
\centering
Soyoung Yoon$^1$\thanks{~~Work done during an internship at Channel Corporation.}\hspace{0.4em}\hspace{1em}
  Dongha Ahn$^1$$^2$\hspace{1em}
  Youngwon Lee$^1$\hspace{1em}
  Minkyu Jung$^2$\hspace{1em}
  HyungJoo Jang$^2$\hspace{1em}
  Seung-won Hwang$^1$\thanks{~~Corresponding author.}
  }\vspace{0.12cm}\\
  $^1$Seoul National University
  $^2$Channel Corporation
\\
\texttt{\{soyoung.yoon, seungwonh\}@snu.ac.kr}
}
\vspace{-0.5cm}

\begin{document}
\maketitle
\begin{abstract}
Mitigating positional bias of language models (LMs) for \textbf{listwise} inputs is a well-known and important problem (e.g., lost-in-the-middle). While zero-shot order-invariant LMs have been proposed to solve this issue, their success on practical listwise problems has been limited. In this work, as a first contribution, we identify and overcome two limitations to make zero-shot invariant LMs more practical: \textbf{(1)} training and inference distribution mismatch arising from modifying positional ID assignments to enforce invariance, and \textbf{(2)} failure to adapt to mixture of order-invariant and sensitive inputs in practical listwise problems. Then, to overcome these issues we propose \textbf{(1)} RoToR, a zero-shot invariant LM for genuinely order-invariant inputs with minimal modifications of positional IDs, and \textbf{(2)} \sr{}, an adaptive framework that handles both order-invariant and order-sensitive inputs in listwise tasks. On the Lost in the middle (LitM), Knowledge Graph QA (KGQA), and MMLU benchmarks, we show that \ours{} with \sr{} can effectively handle practical listwise input tasks in a zero-shot manner.\footnote{\url{https://github.com/soyoung97/RoToR}}
\end{abstract}

\section{Introduction}
\begin{figure}[!t]

\centering
\includegraphics[width=\linewidth]{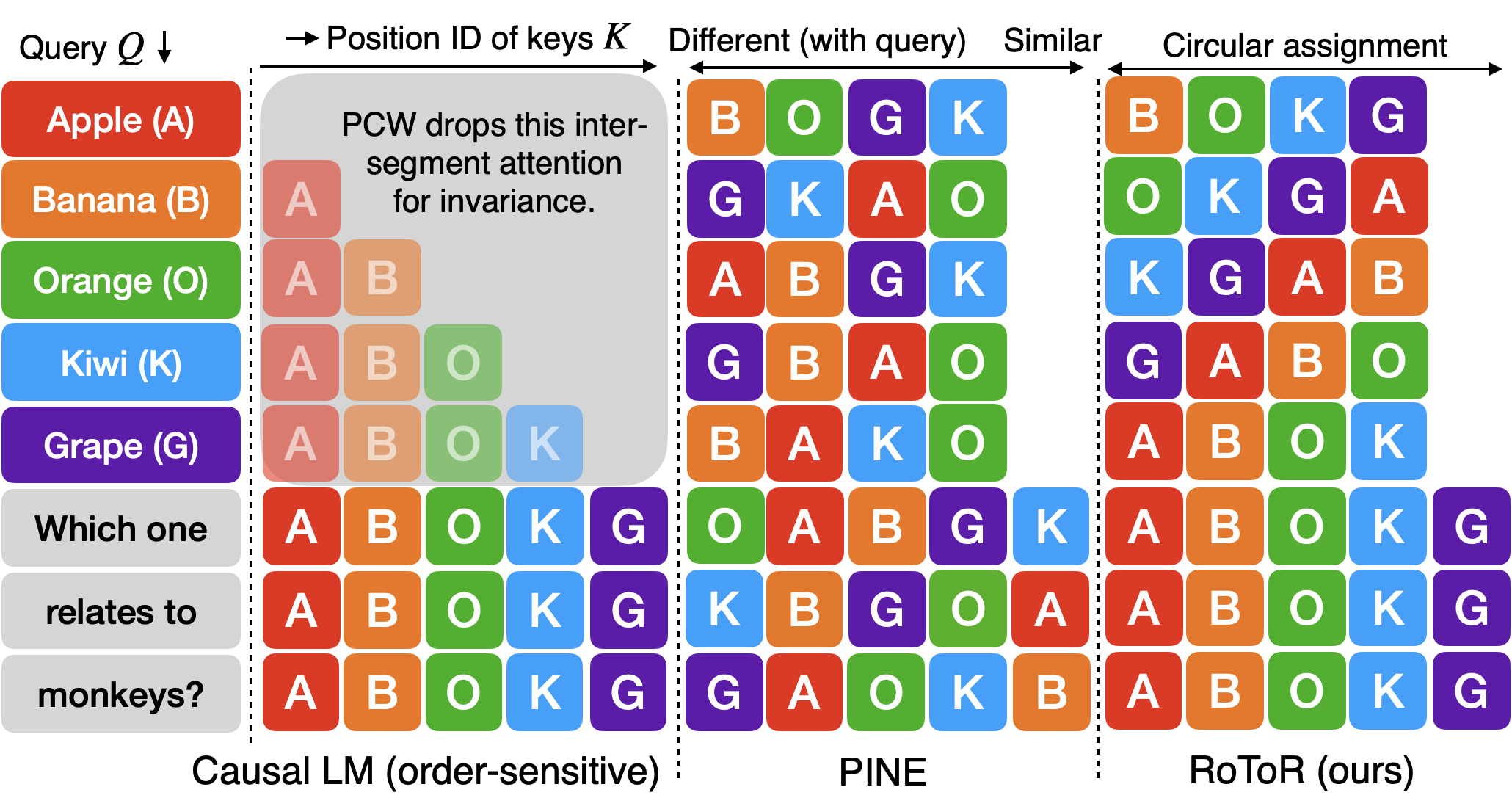}
\caption{Self-attention alteration from  order-invariant models.
(a) PCW by elimination (b)
PINE by
re-assignment of position IDs 
based on
query-based pairwise ordering. In contrast, (c) RoToR minimizes the distribution mismatch by global ordering with circular assignment.}

\label{fig:intro_example}
\end{figure}

Language conveys meaning in part through positional information, such as word placement and sentence structure.
Given this nature, Language Models (LMs) that learn from human language are trained sensitive to positional information related to the ordering of segments. However, there are some listwise inputs that
require neutrality to positional information.
For example, for inputs such as sets, tables, databases, or multiple-choice questions, the ordering of the input \textbf{segments}---e.g., rows in a table or elements in an unordered set---require an order-agnostic understanding. We refer to such inputs as ``order-invariant inputs,'' on which LMs reportedly struggle.
For example, in LLM-as-a-judge scenarios, LMs exhibit a preference of up to 75\% for the first answer in pairwise inputs~\cite{zheng2024judging}, and ranking between LMs can change up to 8 positions in different orderings of multiple choice questions on MMLU~\cite{alzahrani2024benchmarkstargetsrevealingsensitivity}. 
Such results question the reliability of LMs on order-invariant inputs. Meanwhile, existing methods for enforcing invariance to LMs showed limited effectiveness in real-world tasks, which we hypothesize to arise from the following limitations.

\textbf{First,} training and inference distribution mismatch due to the positional ID re-assignment of zero-shot order-invariant LMs: Fig.~\ref{fig:intro_example} illustrates how self-attention is altered in these models. Unlike the original non-invariant model which always assigns position IDs in a causal, ascending manner, order-invariant models either eliminate inter-segment attention, such as PCW~\cite{pcw} in Fig.~\ref{fig:intro_example}a,
or re-assign position IDs as in PINE~\cite{pine} inFig.~\ref{fig:intro_example}b, re-ordering segments using pairwise similarity, placing similar segments closer to the query. For each query segment, it computes segment-wise query-key attention (for each attention head in each decoder layer) and re-assigns position IDs of segments as keys. This query-dependent segment ordering leads to excessively frequent alterations of positional ID assignments. Frequent re-assignments can also confuse the model and risk collisions which violate the invariance property (e.g., multiple key segments having the same similarity to a query).

To overcome this, we propose a query-agnostic global sorting with circular arrangement for order-invariant positional ID assignment. 
Ours is named \textbf{\ours{}}, inspired by the word \textit{rotary} to express circular assignment, and also a palindrome, to reflect order invariance.

Fig.~\ref{fig:intro_example}c contrasts with
 PINE in Fig.~\ref{fig:intro_example}c, where \ours{} only needs a single global ordering (e.g., \texttt{A->B->O->K->G}) with no extra attention computation. The ordering of segments on suffix tokens remains in a fixed order, since it does not rely on their similarity to the query.
Finally, we propose three different global sorting algorithms for \ours{}, and demonstrate that they consistently outperform previous order-invariant models.

\textbf{Second,} for practical listwise inputs, 
order-invariant tasks may partially include
order-sensitive inputs that require order-specific understanding. For example, the \texttt{(d) None of the above} option in MMLU cannot be reordered. Such a ``mixed'' nature requires handling each of the cases adaptively, for which we propose a simple \sr{} method. \sr{} adapts to a given input by routing between two models, invariant and non-invariant (original), based on the confidence scores of their predictions. Experiments on the MMLU benchmark show that \sr{} effectively handles datasets with order-invariant and sensitive inputs, and achieves better order robustness while maintaining the original performance.

In summary, our contributions are as follows:
\textbf{1. Clarifying key challenges to robust understanding of listwise inputs.} We pinpoint the distribution mismatch and positional ID assignment complexities that hinder zero-shot order-invariance in LMs, and the need to adaptively handle order-invariant and order-sensitive inputs. \textbf{2. A stable, order-invariant solution (RoToR):} We propose a query-agnostic global ordering with minimal positional ID modifications, resulting in stable and efficient order-invariance. \textbf{3. Adaptive handling of listwise inputs (\sr{}):} We introduce a simple routing method that switches between the original and invariant LMs based on confidence. On MMLU, we show that \sr{} can adaptively deal with both types of input, leading to better stability.
To this end, we aim to develop a model that excels at processing a wide range of listwise inputs reliably and efficiently.

\section{Related Works}
\subsection{Positional bias of LLMs}
\noindent \textbf{Problem statement.} 
Recent works on (zero-shot) retrieval augmented generation (RAG) with LLMs have found that the models exhibit unwanted bias on the \textit{ordering} of the retrieved documents~\cite{chhabra2024revisitingzeroshotabstractivesummarization}.
Widely known as the lost-in-the-middle problem~\cite{liu2024lost}, many prior studies~\cite{chen2024premise, gupta2024changinganswerorderdecrease, pezeshkpour2023large, zhao2023robut, zhou2024frebtqa, wei2024unveilingselectionbiasesexploring, alzahrani2024benchmarkstargetsrevealingsensitivity, zheng2024large} also investigate the impact of positional bias, extending the domain to 
structured knowledge grounding (SKG) tasks~\cite{zhao2023robut, zhou2024frebtqa} and multiple-choice questions~\cite{gupta2024changinganswerorderdecrease} where changing the ordering of rows, schemas, or choices greatly degrades performance.

\noindent \textbf{Considerations for decoder-only LMs.} While successful approaches are presented to mitigate this issue for encoder-only~\cite{yang2022tableformerrobusttransformermodeling} and encoder-decoder~\cite{yen2024longcontextlanguagemodelingparallel, cai2023scaling} models, they leave decoder-only models, which account for the current frontier LLMs, for more consideration. In contrast to transformer encoders that use bidirectional attention which is invariant by nature~\cite{lee2019set}, transformer decoders use causal attention to learn causal relation signals, which is not invariant by nature~\cite{haviv2022nopos}. Therefore, positional bias for decoder-only models is known to stem from \textit{both} positional encoding and causal attention mask~\cite{yu2024mitigate, pine} and is harder to mitigate.

\subsection{Zero-shot order-invariance for LLMs}
\label{related_work:orderinv}
\textbf{Long context modeling.} Zero-shot approaches for mitigating positional bias in LLMs were first raised in long-context tasks, with a goal to correctly handle relevant information located in the \textit{middle} of lengthy inputs\footnote{\href{https://github.com/gkamradt/LLMTest_NeedleInAHaystack}{\texttt{github.com/gkamradt/LLMTest\_NeedleInAHaystack}}}. Nonetheless, these works focus primarily on understanding long texts without losing precision~\cite{li2023loogle, marathon, leval, longbench}, whereas positional bias is a more general problem that can occur even on multiple-choices questions with relatively short contexts~\cite{alzahrani2024benchmarkstargetsrevealingsensitivity}. Technically, this line of works modify the attention mechanism by altering the positional encoding to adapt an LLM to longer contexts~\cite{peng2023yarn, hsieh2024found, peysakhovich2023attention,chen2023fortify, junqing2023never,xu2023retrieval, yu2024mitigate, zhang2024found}. But since they do not modify the causal mask which also contributes to positional bias, order-invariance is not guaranteed in general~\cite{haviv2022transformerlanguagemodelspositional}.

\noindent \textbf{(Zero-shot) order-invariance.}
Recent line of works focused on achieving order-invariance by mechanistically altering both positional encoding and causal masking.
While several works require training~\cite{junqing2023never, zhu2023judgelm}, we focus on zero-shot approaches for practicality, namely PCW~\cite{pcw}, Set-Based Prompting~\cite{setbasedprompting}, and PINE~\cite{pine}, which we explain in detail at Sec.~\ref{subsec:baseline}.
Another line of works based on self-consistency try to mitigate positional bias simply by running inference multiple times with different orderings of contexts~\cite{zheng2024large}. However, in principle, this requires evaluating \(n!\) forward passes in total, enforcing Monte Carlo approximations~\citep{tang2024middlepermutationselfconsistencyimproves}. More recent work optimizes the number or passes~\citep{lee2024inference} with similar comprehensiveness \citep{hwang2007optimizing}, or replaces with contrastive training objective~\cite{lee2024cord}. In contrast, our method guarantee invariance with a \textit{single} forward pass, without requiring any approximations.

\section{Methodology}
\begin{figure*}[t!]
{
\centering
    \includegraphics[width=0.8\linewidth]{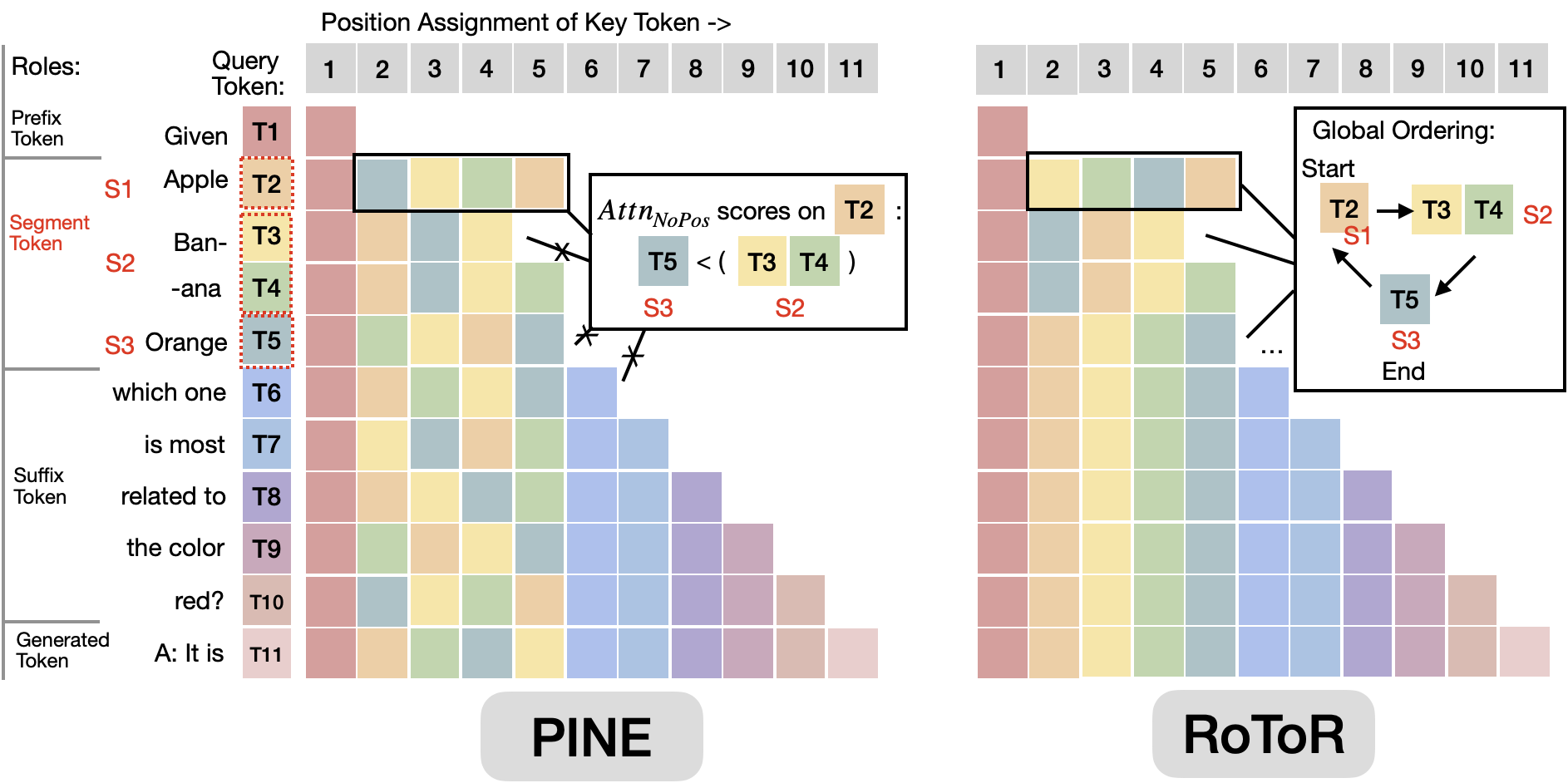}
    \caption{Attention mask and positional ID modifications for segment-wise order invariance using example input \texttt{``Given Apple Banana Orange...''}. Each block represents a single token arranged by positional ID assignment. In practice, tokens retain original positions but receive reassigned positional IDs as shown. White areas indicate masked attention. \textbf{Key difference}: PINE requires per-query sorting while RoToR reuses one global ordering across all queries.}
    \label{fig:fruit_figure}
}
\end{figure*}

\subsection{Baseline: Order-invariant causal LMs}
\label{subsec:baseline}

In this section, we briefly overview the existing work on endowing decoder-only models on order-invariance by adjusting attention mechanism, and review their limitations.

\paragraph{Isolated parallel processing}
Prior works like PCW~\citep{pcw} and Set-Based Prompting~\cite{setbasedprompting} have modified the attention mask and positional ID assignments of the language model to isolate the processing of each segment and apply same positional embeddings are applied across segments, and thus achieve order invariance:
However, this design completely prevents one segment from attending to the others, and aggregating the information from different segments is solely handled at suffix and generated tokens, significantly hindering the LM's cross-segment contextualized understanding of the text.
\citet{yang2023revisitingparallelcontextwindows} have argued that this essentially degenerates to mere ensemble of conditioning on each context separately.
Such information bottleneck and train-test time discrepancy limits the applicability, more severely as the number of segments is increased. 

\paragraph{Bidirectional processing with Q-K similarity}

A more recent work, PINE~\citep{pine} has addressed these issues through a bidirectional attention mechanism that allows each segment to attend to all other segments. To achieve this within decoder-only models while maintaining order invariance, PINE dynamically modifies positional IDs based on whether a token acts as a \textbf{query} or \textbf{key} in the attention computation.

The key insight is that PINE creates an ``illusion'' for each query segment: it assigns the query segment the largest positional IDs among all segments, enabling it to attend to all other segments bidirectionally. The ordering of key segments is then determined by their relevance scores computed without positional embeddings ($\mathrm{Attn}_{\mathrm{NoPoS}}$), ensuring that more relevant segments appear closer to the query.

Consider the example in Fig.~\ref{fig:fruit_figure} with input \texttt{[T1`Given'', S1[Apple''], S2[Ban'', ana''], S3[Orange''], T6`which one'', .. T10`red?'']}. The prefix token \texttt{`Given''} (T1) and suffix tokens \texttt{`which one .. red?''} (T6-T10) maintain their original positions and follow standard causal attention. For the segments S1-S3, PINE applies its order-invariant mechanism:

\textbf{Dynamic positional ID assignment:} When a token from segment S2 (e.g., \texttt{`ana''} at T4) acts as a \textbf{query}, PINE: (1) Assigns S2 the highest positional IDs (4-5) among all segments, placing it last. (2) Maintains internal causal order within S2: \texttt{`ban''} gets position 4, \texttt{`ana''} gets position 5. Then, it (3) Reorders other segments (S1, S3) based on their $\mathrm{Attn}_{\mathrm{NoPoS}}$ scores with S2. Conversely, when the same token acts as a \textbf{key} for another query (e.g., from S1), its position depends on the relevance score between S2 and the query segment. If $\mathrm{Attn}_{\mathrm{NoPoS}}(\text{S1}, \text{S2}) > \mathrm{Attn}_{\mathrm{NoPoS}}(\text{S1}, \text{S3})$, then S2 is placed closer to S1 than S3.

This dynamic reassignment occurs for every attention computation: each query token sees a different positional arrangement of the key tokens, determined by their relevance scores. Prefix, suffix, and generated tokens do not participate in this reordering and always maintain their standard causal positions. However, when these non-segment tokens act as queries, they still see the segments reordered by their relevance scores.

\begin{figure*}[t!]
{
\centering
    \includegraphics[width=\linewidth]{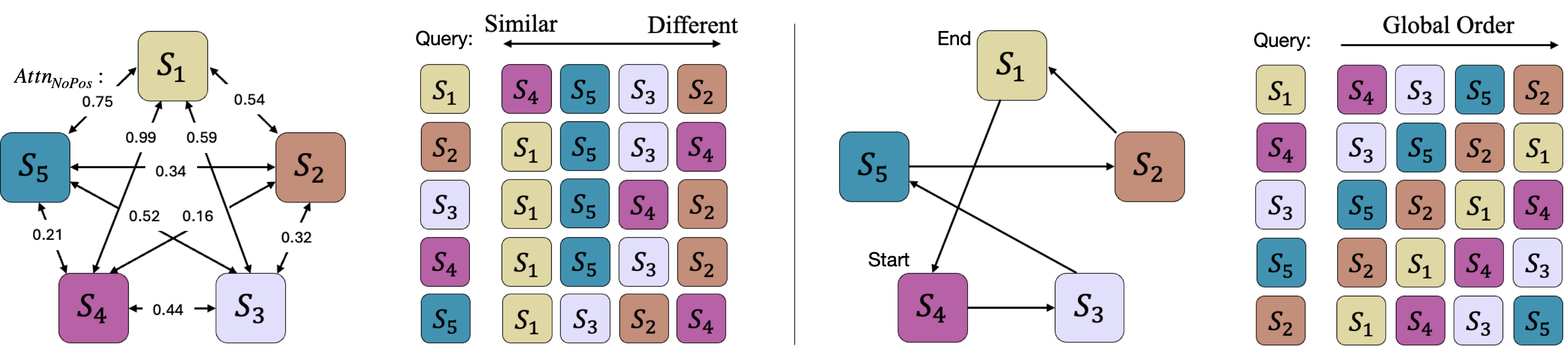}
    \caption{Comparing the ordering of 5 segments (S1 - S5) of PINE~(\citealp{pine}; left) and \ours{} (Ours; right). PINE sorts segments using aggregated attention scores. In order to be fully ordering-invariant, segment sorting is changed per token in suffix level, causing confusion. In contrast, we define one global sorting of segments and conduct circular assignment between segments. With this, we simply use the global sorting for position id assignment on suffix tokens, without harming invariance.}
    \label{fig:ours}
}
\end{figure*}

\subsection{\ours{}: minimal OOD from positional ID assignments}
\label{sec:method_ours}

\paragraph{Challenges with PINE} While PINE achieves order-invariance by contextualization across segments, its query-specific ordering scheme introduces (1) significant train-test behavior discrepancy as well as (2) unnecessary complexity and numerical instability, which limits its scalability. During decoding with PINE, position IDs are assigned differently for every query token (each token in the suffix), decoder layer, and attention head, as the query-key attention score $\mathrm{Attn}_{\mathrm{NoPos}}$ determines the position IDs. This complexity introduces \textbf{excessively frequent alterations on position IDs}: As the base LM is trained with fixed positional IDs and causal masks, this causes hidden activations higher risk of out-of-distribution (OOD) for it to process properly. Moreover, ordering segments based on attention is \textbf{computationally expensive} and introduces \textbf{numerical instability}. 
As computing the attention value of one query segment requires computing the KV attention over every other number of segments, PINE invokes $\mathcal{O}(n^2)$ cost overhead for each segment for input length $n$, which is further multiplied by the number of all combinations of layers, heads, and the number of suffix and generated tokens. 
Also, in practice, calculating attention without RoPE results in a very narrow range of values. \texttt{bfloat16} numeric type lacks precision to distinguish these values, leading to non-determinism originating from several tied values. The outcome may then depend on the initial ordering.

\paragraph{Motivation \& Theoretical Foundation} While investigating ways to overcome the limitations of PINE, our central goal is to preserve order invariance while minimizing the complexity of the re-assignment of positions. We reason that defining a \textit{single} global ordering scheme, not necessarily relying on attention scores, and re-using them across all queries can solve the problems stemming from query-dependent ordering. A circular assignment of a global order seems as a practical solution. The idea of using global sorting to achieve ordering invariance has been studied in set/graph ML domain \citep{murphy2019janossypoolinglearningdeep, murphy2019relationalpoolinggraphrepresentations}, but to the best of our knowledge, using circular assignment of the global ordering, and the application to pre-trained language models, are our novel contributions.
As a result, we propose \ours{}(Fig.~\ref{fig:ours}), which uses one \textbf{global ordering} that is not a function of the initial ordering of segments (e.g., canonical ordering by lexical sorting) and assigns IDs for tokens in different segments based on \textbf{circular arrangement}.

\paragraph{Global ordering}
Instead of re-computing the relative order of segments for each query, we reuse a globally shared single ordering, avoiding costly recomputation of numerically unstable attention scores.
Moreover, this further reduces the gap between the LLM's pretrained behavior and test-time behavior, as consistent position IDs are assigned across layers/heads/across suffix tokens.
Global ordering allows to preserve the relative placement of segments, further closing the gap induced from introducing order invariance to causal LMs.
For example, in Figure~\ref{fig:ours}, due to the global ordering, segments $S_5$ and $S_2$ are always placed in adjacent positions with \ours{} (right side), while it is not satisfied and constantly changed with PINE (left side). 
We consider three separate global sorting algorithm to be used in \ours{}: (1) simple \textbf{lexicographical sorting} which can be obtained with minimal overhead based on tokenized sequence of segments, (2) using a \textbf{pointwise reranker}~\cite{monot5}\footnote{\href{https://huggingface.co/castorini/monot5-base-msmarco-10k}{\sloppy{\texttt{castorini/monot5-base-msmarco-10k}}}} to score relevancy of each row with respect to the question, or (3) simple \textbf{freq}uency-based sorting which normalizes token ids based on the inverse frequency of each token (Details at Appendix Fig.~\ref{fig:global_sort_example}). Empirically, we find that using simple lexicographical sorting is sufficient to obtain improvements over PINE.

\paragraph{Circular arrangement}
To mimic bidirectionality with causal LMs, each segment should be assigned position IDs so that they appear to themselves as being placed at the end of the sequence of segments.
To achieve this with a shared global ordering, we employ circular arrangement, each segment taking turns to be placed at the end while their relative ordering is preserved.
Given the global ordering, we can construct a single directed graph by combining the front and last parts. Then, we assign orderings for each segment as query by following the path from the graph, starting from the query segment, which is illustrated in Fig.~\ref{fig:ours}. For all suffix and generated tokens, segments are arranged according to the initial front and last part of the global ordering. Compared to PINE where we have to assign different orderings of segments for each suffix and generated tokens, \ours{} assign the same positional ID, acting merely the same as the original token. This also accounts for reducing the distributional gap between the original model.

\paragraph{Computational overhead} 
We report only operations executed beyond vanilla self-attention cost $O(n^{2}d)$, where $n$ is total input length, $d$ is hidden dimension, and $k$ is the number of segments.
PINE requires two additional operations: (1) computing attention scores without rotary position embeddings ($\mathcal{O}(n^2d)$) and (2) sorting $k$ segments for each query token ($\mathcal{O}(nk\log k)$), totaling $\boldsymbol{\mathcal{O}(n^{2}d + nk\log k)}$~\citep{pine}\footnote{The PINE paper reports $\mathcal{O}(nk\log k)$ by absorbing the $\mathcal{O}(n^2d)$ term into baseline; we expose it explicitly for fair comparison.}.
In contrast, our lexicographical sorting requires only a single global sort of $k$ segments ($\mathcal{O}(k\log k)$), each with length $\mathcal{O}(n)$, achieving $\boldsymbol{\mathcal{O}(nk\log k)}$ and eliminating the expensive $\mathcal{O}(n^2d)$ term entirely. 
This can be further optimized to $\boldsymbol{\mathcal{O}(nk)}$ using radix sort.\footnote{\url{https://en.wikipedia.org/wiki/Radix_sort}}
We empirically validate significantly faster performance than PINE as $k$ increases (Tab.~\ref{table/inference_cost}).

\subsection{\sr{} for handling order-sensitive inputs}
\label{sec:mov}
\begin{figure}[t!]
{
\centering
    \includegraphics[width=\linewidth]{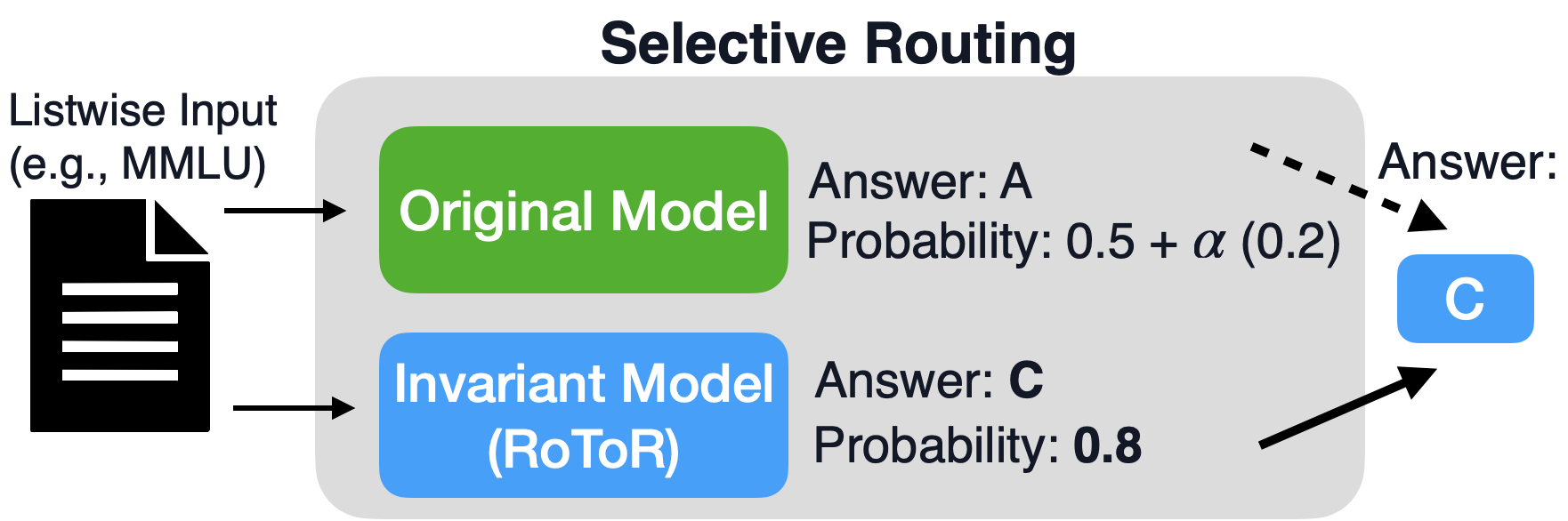}
    \caption{Illustration of \sr{} (Sec.~\ref{appendix:alpha_search}).}
    \label{fig:mov}
}
\end{figure}
Since many practical benchmarks such as MMLU involves semi-invariant inputs, we propose a routing mechanism that uses the order-invariant model in conjunction with the standard causal model for further applicability. Our design is partly based on the finding from
\citet{wei2024unveilingselectionbiasesexploring} that there is correlation between task difficulty (which is in turn correlated with confidence values) and the model's sensitivity to ordering.
\sr{}, illustrated in Fig.~\ref{fig:mov}, combines confidence, the model output probability for the generated answer, from two different model versions---the original model and the order-invariant model---on the same input and choose a more confident answer. 
Both models first produce a maximum probability over possible answer tokens (e.g., \texttt{A, B, C, D} for MMLU) and a corresponding answer choice. We then compare the original model’s maximum probability, plus a bias term $\alpha$, to the invariant model’s maximum probability. If the original model’s adjusted score is higher, we take its answer; otherwise, the invariant model’s answer is chosen.
$\alpha$ is a hyperparameter that controls how strongly the original model is favored, which was selected as 0.2 according to hyperparameter search on the validation subset (Appendix Sec.~\ref{appendix:alpha_search}).

\section{Experiment setup}
\begin{table*}[!t]
\centering
\resizebox{0.85\linewidth}{!}
{
\begin{tabular}{@{}lccccccccccccccc@{}}
\toprule
\multicolumn{1}{l|}{Total ndoc (segments)} & \multicolumn{3}{c|}{10} & \multicolumn{5}{c|}{20} & \multicolumn{7}{c}{30} \\ \midrule
\multicolumn{1}{l|}{Gold idx at:} & 0 & 4 & \multicolumn{1}{c|}{9} & 0 & 4 & 9 & 14 & \multicolumn{1}{c|}{19} & 0 & 4 & 9 & 14 & 19 & 24 & 29 \\ \midrule
\multicolumn{16}{l}{Llama-3.1-8B-Instruct} \\ \midrule
\multicolumn{1}{l|}{Original} & 54.7 & 53.0 & \multicolumn{1}{c|}{50.2} & 54.8 & 52.6 & 52.8 & 52.4 & \multicolumn{1}{c|}{51.0} & 55.6 & 51.5 & 52.4 & 52.8 & 52.1 & 52.3 & 53.0 \\
\multicolumn{1}{l|}{PCW} & 12.4 & 11.9 & \multicolumn{1}{c|}{12.2} & 3.7 & 4.0 & 4.0 & 4.0 & \multicolumn{1}{c|}{3.9} & 2.3 & 1.8 & 2.0 & 2.0 & 2.1 & 2.0 & 2.0 \\
\multicolumn{1}{l|}{Set-Based Prompting} & 42.5 & 42.5 & \multicolumn{1}{c|}{42.5} & 26.3 & 26.3 & 26.3 & 26.3 & \multicolumn{1}{c|}{26.3} & 14.1 & 14.1 & 14.1 & 14.1 & 14.1 & 14.1 & 14.1 \\
\multicolumn{1}{l|}{PINE} & 58.6 & 58.8 & \multicolumn{1}{c|}{59.0} & 56.2 & 55.7 & 55.5 & 55.7 & \multicolumn{1}{c|}{55.5} & 54.2 & 54.8 & 54.3 & 53.7 & 54.8 & 54.2 & 54.0 \\
\multicolumn{1}{l|}{\ours{}-lexical} & 61.4 & 61.6 & \multicolumn{1}{c|}{61.6} & \textbf{61.4} & 59.8 & 59.6 & 59.6 & \multicolumn{1}{c|}{59.8} & 59.2 & 59.5 & 59.4 & 59.1 & 59.0 & 59.3 & 59.1 \\
\multicolumn{1}{l|}{\ours{}-reversed lexical} & \textbf{61.6} & \textbf{61.8} & \multicolumn{1}{c|}{\textbf{61.8}} & 58.9 & 59.3 & 58.8 & 58.6 & \multicolumn{1}{c|}{58.7} & 57.9 & 58.2 & 57.9 & 57.4 & 57.9 & 57.6 & 57.5 \\
\multicolumn{1}{l|}{\ours{}-MonoT5} & 61.2 & 61.4 & \multicolumn{1}{c|}{61.2} & 60.9 & \textbf{61.0} & \textbf{61.2} & \textbf{61.2} & \multicolumn{1}{c|}{\textbf{61.2}} & \textbf{60.9} & \textbf{60.7} & \textbf{60.7} & \textbf{60.7} & \textbf{60.8} & \textbf{60.8} & \textbf{60.7} \\
\multicolumn{1}{l|}{\ours{}-Freq.} & 61.0 & 61.1 & \multicolumn{1}{c|}{61.1} & 60.4 & 60.3 & 58.6 & 60.2 & \multicolumn{1}{c|}{60.0} & 59.3 & 60.4 & 59.7 & 59.5 & 59.5 & 59.6 & 59.2 \\ \midrule
\multicolumn{16}{l}{Llama 3.1-70B-Instruct} \\ \midrule
\multicolumn{1}{l|}{Original}        & 66.2 & 65.7 & \multicolumn{1}{c|}{65.7} & 65.2 & 64.3 & 65.0 & 66.2 & \multicolumn{1}{c|}{64.8} & \multicolumn{7}{c}{--} \\
\multicolumn{1}{l|}{PINE}            & 67.9 & 67.8 & \multicolumn{1}{c|}{67.5} & 65.9 & 65.7 & 65.9 & 65.8 & \multicolumn{1}{c|}{65.5} & \multicolumn{7}{c}{--}\\
\multicolumn{1}{l|}{\ours}           & \textbf{69.6} & \textbf{69.5} & \multicolumn{1}{c|}{\textbf{69.3}} & \textbf{67.6} & \textbf{67.8} & \textbf{67.8} & \textbf{67.7} & \multicolumn{1}{c|}{\textbf{67.9}} & \multicolumn{7}{c}{--} \\
\multicolumn{1}{l|}{\ours{}-MonoT5}  & 68.9 & 69.0 & \multicolumn{1}{c|}{68.8} & 67.5 & 67.5 & 67.7 & 67.5 & \multicolumn{1}{c|}{67.6} & \multicolumn{7}{c}{--} \\
\midrule

\multicolumn{16}{l}{Qwen1.5-4B-Chat} \\ \midrule
\multicolumn{1}{l|}{Original} & 61.3 & 54.8 & \multicolumn{1}{c|}{53.1} & \textbf{59.5} & 49.1 & 47.9 & 45.9 & \multicolumn{1}{c|}{48.3} & \textbf{56.8} & 45.6 & 44.9 & 44.6 & 45.3 & 43.5 & 48.3 \\
\multicolumn{1}{l|}{PINE} & 57.2 & 57.4 & \multicolumn{1}{c|}{57.0} & 48.6 & 48.2 & 48.2 & 48.1 & \multicolumn{1}{c|}{48.9} & 46.4 & 45.9 & 46.7 & 46.6 & 46.4 & 46.4 & 46.3 \\
\multicolumn{1}{l|}{\ours} & 58.5 & 58.4 & \multicolumn{1}{c|}{58.1} & 49.9 & 49.7 & 49.6 & 49.8 & \multicolumn{1}{c|}{49.9} & 44.6 & 44.8 & 44.7 & 44.7 & 44.9 & 44.8 & 44.7 \\
\multicolumn{1}{l|}{\ours{}-MonoT5} & \textbf{58.9} & \textbf{58.5} & \multicolumn{1}{c|}{\textbf{58.7}} & 52.2 & \textbf{52.1} & \textbf{52.1} & \textbf{52.2} & \multicolumn{1}{c|}{\textbf{52.6}} & 50.6 & \textbf{50.7} & \textbf{50.5} & \textbf{50.6} & \textbf{50.5} & \textbf{50.6} & \textbf{50.4} \\
\multicolumn{1}{l|}{\ours{}-Freq.} & 56.7 & 56.9 & \multicolumn{1}{c|}{56.9} & 51.9 & 51.5 & 51.8 & 51.6 & \multicolumn{1}{c|}{52.4} & 46.8 & 46.7 & 46.7 & 46.4 & 47.0 & 46.8 & 46.6 \\ \midrule
\multicolumn{16}{l}{Qwen1.5-7B-Chat} \\ \midrule
\multicolumn{1}{l|}{Original} & \textbf{72.5} & 63.3 & \multicolumn{1}{c|}{62.9} & \textbf{72.5} & 58.5 & 56.1 & 56.0 & \multicolumn{1}{c|}{58.2} & \textbf{73.1} & 58.6 & 55.8 & 53.3 & 53.2 & 52.5 & 57.5 \\
\multicolumn{1}{l|}{PINE} & 65.4 & 65.5 & \multicolumn{1}{c|}{66.3} & 59.1 & 59.4 & 59.1 & 58.6 & \multicolumn{1}{c|}{59.2} & 58.0 & 55.3 & 55.7 & 56.3 & 55.1 & 55.8 & 56.1 \\
\multicolumn{1}{l|}{\ours} & 68.6 & 68.7 & \multicolumn{1}{c|}{68.6} & 62.6 & 62.9 & 62.7 & 63.0 & \multicolumn{1}{c|}{62.7} & 57.0 & 57.3 & 59.7 & 57.4 & 57.3 & \textbf{62.8} & 57.0 \\
\multicolumn{1}{l|}{\ours{}-MonoT5} & 68.8 & \textbf{69.4} & \multicolumn{1}{c|}{\textbf{69.0}} & 65.2 & \textbf{65.5} & \textbf{65.0} & \textbf{64.9} & \multicolumn{1}{c|}{\textbf{65.0}} & 62.6 & \textbf{62.8} & \textbf{62.9} & \textbf{62.7} & \textbf{62.9} & \textbf{62.8} & \textbf{62.5} \\
\multicolumn{1}{l|}{\ours{}-Freq.} & 68.2 & 68.4 & \multicolumn{1}{c|}{68.4} & 62.6 & 62.9 & 62.8 & 62.7 & \multicolumn{1}{c|}{62.3} & 59.5 & 59.8 & 59.7 & 59.6 & 59.7 & 59.7 & 59.7 \\ \bottomrule
\end{tabular}
}

\caption{
The \texttt{best\_subspan\_em} ($\%$) scores on the \textbf{lost in the middle (LitM)} benchmark, with indexing bias removed,  across varying numbers of documents (ndoc $\in \{10, 20, 30\}$) and models. \ours{} shows the best performance across all setups. Experiments on ndoc=30 for the Llama 70B model were unable to report due to resource constraints.
}
\label{table/litm_number}

\end{table*}

\newcommand{\best}[1]{\textbf{#1}}

\begingroup

  \renewcommand{\arraystretch}{0.95}

  \setlength{\extrarowheight}{-1pt}
\begin{table*}[t]
\centering
\resizebox{0.93\linewidth}{!}{
\begin{tabular}{@{}l|ccc|ccc|ccc|ccc|ccc@{}}
\toprule
\multicolumn{1}{l|}{} &
\multicolumn{3}{c|}{Llama-3.1-8B-Instr.} &
\multicolumn{3}{c|}{Llama-3.1-70B-Instr.} &
\multicolumn{3}{c|}{Qwen1.5-4B-Chat} &
\multicolumn{3}{c|}{Qwen1.5-7B-Chat} &
\multicolumn{3}{c}{Qwen1.5-72B-Chat} \\ \cmidrule(l){2-16}
\multicolumn{1}{l|}{\textbf{Method}} & Acc. & EM & F1 & Acc. & EM & F1 & Acc. & EM & F1 & Acc. & EM & F1 & Acc. & EM & F1 \\ \midrule
\rowcolor{yellow!20}\multicolumn{16}{@{}>{\raggedright\arraybackslash}c}{\boldsymbol{$N\!=\!30$}} \\ \midrule
\multicolumn{16}{l}{\textbf{Initial, no shuffling of segments}} \\ \midrule
Original        & 50.2 & 44.0 & 51.9 & 61.6 & 57.7 & 63.6 & 30.7 & 27.9 & 34.9 & 31.5 & 27.8 & 35.4 & 41.4 & 37.7 & 43.7 \\
PINE            & 51.5 & 45.0 & 52.6 & 63.1 & 58.7 & 64.8 & 31.6 & 28.7 & 35.6 & 32.3 & 28.8 & 36.4 & 46.7 & 42.9 & 49.0 \\
RoToR           & \best{53.1} & \best{46.5} & \best{54.1} & \best{63.6} & \best{59.1} & \best{65.2} & 32.0 & 29.0 & 35.7 & \best{34.3} & \best{29.8} & \best{37.7} & \best{47.5} & \best{43.2} & \best{49.2} \\
RoToR-MonoT5    & 51.6 & 45.0 & 52.5 & \multicolumn{3}{c|}{--} & \best{32.3} & 29.1 & \best{36.2} & 32.9 & 28.4 & 36.3 & \multicolumn{3}{c}{--} \\
RoToR-Freq.     & 52.6 & 46.1 & 53.7 & \multicolumn{3}{c|}{--} & \best{32.3} & \best{29.2} & 36.0 & 33.7 & 29.5 & 37.2 & \multicolumn{3}{c}{--} \\ \midrule
\multicolumn{16}{l}{\textbf{After shuffling segments, averaged over 3 seeds}} \\ \midrule
Original        & 49.5 & 43.3 & 51.0 & 62.1 & 57.8 & 64.0 & 30.1 & 27.5 & 34.7 & 31.4 & 27.3 & 35.0 & 41.0 & 37.6 & 43.6 \\
\multicolumn{1}{r|}{$\hookrightarrow$ stdev. ($\pm$)} &
\multicolumn{3}{c|}{\small 0.07 / 0.14 / 0.17} &
\multicolumn{3}{c|}{\small 0.37 / 0.40 / 0.27} &
\multicolumn{3}{c|}{\small 0.41 / 0.34 / 0.43} &
\multicolumn{3}{c|}{\small 0.26 / 0.28 / 0.29} &
\multicolumn{3}{c}{\small 0.75 / 0.40 / 0.33} \\
PINE            & 51.8 & 45.2 & 52.8 & 63.3 & 58.8 & 64.9 & 31.5 & 28.7 & 35.6 & 32.3 & 28.8 & 35.7 & 46.9 & \best{43.3} & \best{49.2} \\
\multicolumn{1}{r|}{$\hookrightarrow$ stdev. ($\pm$)} &
\multicolumn{3}{c|}{\small 0.05 / 0.07 / 0.16} &
\multicolumn{3}{c|}{\small 0.13 / 0.04 / 0.10} &
\multicolumn{3}{c|}{\small 0.20 / 0.18 / 0.13} &
\multicolumn{3}{c|}{\small 0.17 / 0.20 / 0.13} &
\multicolumn{3}{c}{\small 0.18 / 0.20 / 0.20} \\
RoToR           & \best{52.8} & \best{46.2} & \best{53.8} & \best{63.5} & \best{59.1} & \best{65.3} & 31.8 & 28.8 & 35.5 & \best{34.2} & \best{29.9} & \best{37.7} & \best{47.4} & 43.1 & 49.1 \\
\multicolumn{1}{r|}{$\hookrightarrow$ stdev. ($\pm$)} &
\multicolumn{3}{c|}{\small 0.05 / 0.05 / 0.02} &
\multicolumn{3}{c|}{\small 0.11 / 0.07 / 0.08} &
\multicolumn{3}{c|}{\small 0.05 / 0.02 / 0.09} &
\multicolumn{3}{c|}{\small 0.09 / 0.07 / 0.06} &
\multicolumn{3}{c}{\small 0.06 / 0.04 / 0.07} \\
RoToR-MonoT5    & 51.6 & 45.0 & 52.6 & \multicolumn{3}{c|}{--} & \best{32.4} & 29.2 & \best{36.3} & 33.0 & 28.8 & 36.5 & \multicolumn{3}{c}{--} \\
\multicolumn{1}{r|}{$\hookrightarrow$ stdev. ($\pm$)} &
\multicolumn{3}{c|}{\small 0.12 / 0.06 / 0.10} &
\multicolumn{3}{c|}{--} &
\multicolumn{3}{c|}{\small 0.04 / 0.02 / 0.13} &
\multicolumn{3}{c|}{\small 0.12 / 0.09 / 0.07} &
\multicolumn{3}{c}{--} \\
RoToR-Freq.     & 52.5 & 45.9 & 53.5 & \multicolumn{3}{c|}{--} & 32.3 & \best{29.3} & 36.0 & 33.8 & 29.6 & 37.4 & \multicolumn{3}{c}{--} \\
\multicolumn{1}{r|}{$\hookrightarrow$ stdev. ($\pm$)} &
\multicolumn{3}{c|}{\small 0.10 / 0.15 / 0.11} &
\multicolumn{3}{c|}{--} &
\multicolumn{3}{c|}{\small 0.13 / 0.16 / 0.09} &
\multicolumn{3}{c|}{\small 0.04 / 0.00 / 0.09} &
\multicolumn{3}{c}{--} \\ \midrule[0.7pt]
\rowcolor{yellow!20}\multicolumn{16}{@{}>{\raggedright\arraybackslash}c}{\boldsymbol{$N\!=\!50$}} \\ \midrule
\multicolumn{16}{l}{\textbf{Initial, no shuffling of segments}} \\ \midrule
Original        & 50.0 & 44.0 & 51.7 & 62.6 & 58.5 & 64.5 & 31.6 & 28.6 & 35.8 & 31.7 & 28.0 & 35.7 & 42.1 & 38.7 & 44.5 \\
PINE            & 51.6 & 45.1 & 52.6 & 64.1 & 59.8 & 65.8 & 31.6 & 28.8 & 35.3 & 32.0 & 28.5 & 35.9 & 48.0 & 44.1 & 49.9 \\
RoToR           & 52.9 & 46.0 & 53.6 & \best{64.6} & \best{60.0} & \best{66.2} & \best{32.7} & \best{29.6} & \best{36.2} & \best{34.3} & \best{30.1} & \best{38.0} & \best{48.4} & \best{44.3} & \best{50.3} \\
RoToR-MonoT5    & 52.4 & 45.4 & 52.8 & \multicolumn{3}{c|}{--} & 32.3 & 29.3 & 35.9 & 32.9 & 28.9 & 36.6 & \multicolumn{3}{c}{--} \\
RoToR-Freq.     & \best{53.1} & \best{46.4} & \best{53.7} & \multicolumn{3}{c|}{--} & 32.3 & 29.2 & 36.1 & 33.5 & 29.5 & 37.2 & \multicolumn{3}{c}{--} \\ \midrule
\multicolumn{16}{l}{\textbf{After shuffling segments, averaged over 3 seeds}} \\ \midrule
Original        & 49.7 & 43.5 & 51.0 & 62.8 & 58.5 & 64.5 & 30.3 & 27.6 & 35.0 & 31.6 & 27.9 & 35.5 & 42.1 & 38.9 & 44.7 \\
\multicolumn{1}{r|}{$\hookrightarrow$ stdev. ($\pm$)} &
\multicolumn{3}{c|}{\small 0.34 / 0.28 / 0.46} &
\multicolumn{3}{c|}{\small 0.29 / 0.28 / 0.05} &
\multicolumn{3}{c|}{\small 0.26 / 0.24 / 0.35} &
\multicolumn{3}{c|}{\small 0.40 / 0.56 / 0.42} &
\multicolumn{3}{c}{\small 0.30 / 0.40 / 0.35} \\
PINE            & 51.8 & 45.3 & 52.7 & 64.3 & 59.8 & 65.9 & 31.5 & 28.7 & 35.3 & 31.7 & 28.2 & 35.7 & \best{48.0} & \best{44.3} & 50.0 \\
\multicolumn{1}{r|}{$\hookrightarrow$ stdev. ($\pm$)} &
\multicolumn{3}{c|}{\small 0.15 / 0.16 / 0.19} &
\multicolumn{3}{c|}{\small 0.16 / 0.15 / 0.14} &
\multicolumn{3}{c|}{\small 0.17 / 0.20 / 0.21} &
\multicolumn{3}{c|}{\small 0.18 / 0.16 / 0.14} &
\multicolumn{3}{c}{\small 0.02 / 0.04 / 0.05} \\
RoToR           & 52.7 & 45.9 & 53.5 & \best{64.5} & \best{60.0} & \best{66.1} & \best{32.5} & \best{29.6} & \best{36.1} & \best{34.2} & \best{30.1} & \best{38.0} & \best{48.3} & \best{44.3} & \best{50.3} \\
\multicolumn{1}{r|}{$\hookrightarrow$ stdev. ($\pm$)} &
\multicolumn{3}{c|}{\small 0.05 / 0.09 / 0.04} &
\multicolumn{3}{c|}{\small 0.02 / 0.02 / 0.01} &
\multicolumn{3}{c|}{\small 0.11 / 0.06 / 0.09} &
\multicolumn{3}{c|}{\small 0.06 / 0.05 / 0.04} &
\multicolumn{3}{c}{\small 0.05 / 0.09 / 0.05} \\
RoToR-MonoT5    & 52.2 & 45.2 & 52.8 & \multicolumn{3}{c|}{--} & 32.3 & 29.4 & 35.9 & 32.8 & 28.8 & 36.5 & \multicolumn{3}{c}{--} \\
\multicolumn{1}{r|}{$\hookrightarrow$ stdev. ($\pm$)} &
\multicolumn{3}{c|}{\small 0.16 / 0.18 / 0.18} &
\multicolumn{3}{c|}{--} &
\multicolumn{3}{c|}{\small 0.16 / 0.13 / 0.07} &
\multicolumn{3}{c|}{\small 0.16 / 0.09 / 0.07} &
\multicolumn{3}{c}{--} \\
RoToR-Freq.     & \best{53.1} & \best{46.4} & \best{53.7} & \multicolumn{3}{c|}{--} & 32.4 & 29.3 & \best{36.1} & 33.7 & 29.6 & 37.4 & \multicolumn{3}{c}{--} \\
\multicolumn{1}{r|}{$\hookrightarrow$ stdev. ($\pm$)} &
\multicolumn{3}{c|}{\small 0.02 / 0.07 / 0.03} &
\multicolumn{3}{c|}{--} &
\multicolumn{3}{c|}{\small 0.09 / 0.04 / 0.06} &
\multicolumn{3}{c|}{\small 0.04 / 0.16 / 0.22} &
\multicolumn{3}{c}{--} \\ \bottomrule
\end{tabular}}
\caption{\textbf{Mintaka (KGQA)} results, with \textbf{Initial} and \textbf{After-shuffle} settings, across different model parameter size and backbones. $N$ refers to the number of top-k segments per query. Rows with “$\hookrightarrow$ stdev.” report standard deviation over 3 seeds.}
\label{table/kgqa}
\end{table*}

\endgroup

\begin{table*}[t]
\centering
\resizebox{0.8\linewidth}{!}
{
\begin{tabular}{l|ccc|ccc|ccc}
\hline
 & \multicolumn{3}{c|}{\textbf{Llama-3.1-8B-Instruct}} & \multicolumn{3}{c|}{\textbf{Qwen1.5-4B-Chat}} & \multicolumn{3}{c}{\textbf{Qwen1.5-7B-Chat}} \\ \hline
\textbf{Method} & \textbf{Init.} & \textbf{Rev.} & \textbf{Avg.} & \textbf{Init.} & \textbf{Rev.} & \textbf{Avg.} & \textbf{Init.} & \textbf{Rev.} & \textbf{Avg.} \\ \hline
\textbf{Orig.} & 68.3 & 64.8 & 65.5 $\pm$ 1.0 & 53.6 & 51.9 & 52.6 $\pm$ 0.6 & \textbf{\textbf{60.1}} & 56.6 & 58.6 $\pm$ 0.9 \\
\textbf{PCW} & 57.0 & 55.1 & 56.1 $\pm$ 1.1 & \multicolumn{3}{c|}{--} & \multicolumn{3}{c}{--} \\
\textbf{Set-Based Prompting} & 31.1 & 33.0 & 31.6 $\pm$ 0.8 & \multicolumn{3}{c|}{--} & \multicolumn{3}{c}{--} \\
\textbf{PINE} & 64.8 & 63.3 & 63.6 $\pm$ 0.7 & 50.5 & 49.3 & 49.4 $\pm$ 0.5 & 57.0 & 54.4 & 55.8 $\pm$ 0.9 \\
\textbf{RoToR} & 63.2 & 62.6 & 62.8 $\pm$ 0.7 & 49.6 & 47.7 & 48.3 $\pm$ 0.7 & 56.5 & 55.8 & 56.2 $\pm$ 0.6 \\
$\hookrightarrow$ + S.R. & \textbf{68.5} & 65.1 & 65.7 $\pm$ 0.9 & 53.7 & 51.8 & \textbf{52.6 $\pm$ 0.6} & \textbf{60.1} & \textbf{57.4} & \textbf{58.8 $\pm$ 0.7} \\
\textbf{RoToR - MonoT5} & 64.2 & 62.9 & 63.5 $\pm$ 0.5 & 49.7 & 47.6 & 48.7 $\pm$ 0.7 & 56.2 & 54.4 & 55.5 $\pm$ 0.7 \\
$\hookrightarrow$ + S.R. & 68.4 & 65.2 & 65.8 $\pm$ 0.9 & \textbf{53.8} & 51.9 & \textbf{52.6 $\pm$ 0.6} & \textbf{60.1} & 57.3 & 58.7 $\pm$ 0.8 \\
\textbf{RoToR - Freq.} & 64.3 & 63.6 & 63.8 $\pm$ 0.6 & 49.9 & 47.6 & 48.7 $\pm$ 0.5 & 56.4 & 54.7 & 55.7 $\pm$ 0.7 \\
$\hookrightarrow$ + S.R. & \textbf{68.5} & \textbf{65.3} & \textbf{65.8 $\pm$ 0.8} & 53.7 & \textbf{52.3} & \textbf{52.6 $\pm$ 0.6} & 60.0 & 57.3 & 58.6 $\pm$ 0.8 \\
\textbf{RoToR + S.R. (Oracle)} & 75.0 & 71.9 & 72.7 $\pm$ 1.0 & 61.8 & 60.1 & 61.1 $\pm$ 1.0 & 68.1 & 66.2 & 67.2 $\pm$ 0.7 \\ \hline
\end{tabular}
}
\caption{Improving applicability to general listwise tasks (\textbf{MMLU}, N=4) with \textbf{\sr{}} (S.R), which includes \textbf{both} order-invariant \textbf{and} order-sensitive examples. Init./Rev. refer to original/reversed orderings, Avg. is the average selection ratio across all (4!-1) re-orderings with standard deviation. S.R (Oracle) represents the upper bound with perfect routing accuracy. \ours{} with \sr{} shows improved performance and stability across input re-orderings.}

\label{table/mmlu}

\end{table*}

\subsection{Baselines}
\label{sec:ablations}
Original causal LM with no modifications (Orig.) were compared, which processes text sequentially. Also, we compare \ours{} against previous zero-shot order-invariant LMs discussed in Sec.~\ref{subsec:baseline}, namely PCW~\cite{pcw}, PINE~\cite{pine}, and Set-Based Prompting~\cite{setbasedprompting}.
We use the LLaMA 3.1~\cite{llama3} 8B-Instruct\footnote{\texttt{meta-llama/Meta-Llama-3.1-8B-Instruct}} 70B-Instruct, Qwen1.5-4B-Chat, Qwen1.5-7B-Chat\footnote{\texttt{Qwen/Qwen1.5-4/7B-Chat}}, and Qwen1.5-72B-Chat as our backbone model for our experiments. As our method \textbf{doesn't need training}, a single A6000 GPU was sufficient to run all of the experiments except for the Llama-3.1-70B-Instruct and Qwen1.5-72B-Chat model.
We also conduct experiments on a subset of benchmarks (LitM and MMLU) on the runtime latency, perplexity, and collision rate of PINE and \ours{}, to further validate our claims on Sec.~\ref{sec:method_ours}.

\subsection{Benchmarks with listwise inputs}\label{sec:benchmarks}

We experiment with three benchmarks involving real-world listwise input data.
Examples of exact inputs and outputs are provided in Appendix~\ref{appendix:inputexamples}.
All reported scores are rounded to the nearest tenth, except for the standard deviation (rounded to the second decimal place). 

\noindent \textbf{Knowledge Graph QA (KGQA)} In KGQA tasks, the model takes facts over knowledge graphs represented as (subject, relation, object), and answers the given question based on the given facts. We basically follow the KGQA dataset preprocessing and evaluation setup from ~\citet{baek2023knowledgeaugmentedlanguagemodelverification}, which uses Mintaka~\cite{mintaka} with Wikidata for knowledge source, and use the Exact Match (EM), Accuracy (Acc), and F1 score metric for evaluation. We also use MPNet~\cite{song2020mpnetmaskedpermutedpretraining} as a dense retriever to retrieve top-k facts over each question, and experiment with segment size of 30 and 50. Replication details and example dataset format are at Appendix Sec.~\ref{appendix:dataset_details} and Fig.~\ref{Mintaka}. Along with measuring the performance of the initial input ordering, we report performance after we \textbf{shuffle} the order of segments with 3 different seeds to see shuffle robustness.

\noindent \textbf{Lost in the middle (LitM)}
We use the Lost in the Middle (LitM) benchmark~\cite{liu2024lost}, which draws from 2655 queries in the Natural Questions (NQ) dataset. It provides sets of (10, 20, 30) passages, placing the gold passage at predetermined positions (e.g., 0, 4, 9) and filling the remaining slots with irrelevant passages.
Following ~\citet{liu2024lost}, the \texttt{best\_subspan\_em} metric is used. Experiments on LitM found that eliminating the effect of index bias is another important detail for measuring true order robustness: (Appendix Sec.~\ref{appendix:index_bias}). Thus, we report experiments with index bias eliminated. The exact prompt and full results including index bias is reported at Appendix Fig.~\ref{lostinthemiddlenoindexing} and Sec.~\ref{appendix:litm_detail}.

\noindent \textbf{MMLU}
The Massive Multitask Language Understanding (MMLU) benchmark~\cite{hendrycks2021measuringmassivemultitasklanguage} (prompts at Appendix Fig.~\ref{mmlu_prompt}) consists of 57 diverse sub-tasks with a total of 14,015 queries to measure general performance of LMs about the knowledge of the world. Despite its popularity, many works report that performance fluctuates heavily depending on the order of choices~\cite{gupta2024changinganswerorderdecrease, pezeshkpour2023large, wei2024unveilingselectionbiasesexploring, alzahrani2024benchmarkstargetsrevealingsensitivity, zheng2024large} and is widely investigated to measure the positional bias of the model. We notice that a lot of proportions consist of ordering-sensitive inputs, which showed the effectiveness of adaptively applying \sr{}.
We additionally report the average performance for all possible (4!-1) re-orderings.

\section{Results \& Analysis}

We report results for KGQA in Tab.~\ref{table/kgqa}, and results for MMLU in Tab.~\ref{table/mmlu}. 
Results for LitM are in Tab.~\ref{table/litm_number}, with a visualization in Appendix Fig.~\ref{fig:litm}.
We use lexical sorting for \ours{} unless stated otherwise.

\paragraph{Effectiveness of \ours{}}
We observe that shuffling input segments leads to non-trivial performance degradations in the original model, which exhibits a statistically significant performance drop on our experimented dataset (two-tailed t-test, p < 0.05, Appendix~\ref{appendix:statistical_significance}). In contrast, our proposed RoToR model does not show a statistically significant difference in performance before and after shuffling, indicating that it is more robust against such perturbations.
On LitM (Tab.~\ref{table/litm_number}), we notice PCW and Set-Based Prompting has impractical performance, with PINE degrading heavily as number of documents ($k$) increases, while RoToR is less affected. On KGQA (Tab.~\ref{table/kgqa}), we show \ours{} outperform PINE with lower standard deviation across shuffled segments, consistent with different model architectures.

\paragraph{Improvements from PINE}
\begin{table}[t]
\centering
\resizebox{0.95\linewidth}{!}{
\begin{tabular}{@{}l l|ccc@{}}
\toprule
\textbf{Model} & \textbf{Benchmark} & \textbf{PINE} & \textbf{RoToR} & \textbf{Reduction} \\
\midrule
\rowcolor{yellow!20}\multicolumn{5}{l}{\textbf{(a) Overhead FLOPs, relative to original model}} \\
\midrule
\multirow{3}{*}{\makecell[l]{Llama-3.1-\\8B-Instruct}}
    & MMLU, $N=4$          & 0.59$\times$  & \textbf{0.55$\times$} & 7.6\%  \\
    & LitM, $N=10$         & 7.07$\times$  & \textbf{4.81$\times$} & 31.9\% \\
    & LitM, $N=30$         & 22.43$\times$ & \textbf{15.05$\times$}& 32.9\% \\
\midrule
\multirow{2}{*}{\makecell[l]{Llama-3.1-\\70B-Instruct}}
    & KGQA, $N=30$         & 1.27$\times$  & \textbf{0.94$\times$} & 26.0\%  \\
    & KGQA, $N=50$         & 1.82$\times$  & \textbf{1.29$\times$} & 29.0\%  \\
\midrule
\multirow{2}{*}{\makecell[l]{Qwen1.5-\\72B-Chat}}
    & KGQA, $N=30$         & 0.45$\times$  & \textbf{0.01$\times$} & 98.0\%  \\
    & KGQA, $N=50$         & 0.58$\times$  & \textbf{0.03$\times$} & 94.8\%  \\
\midrule
\rowcolor{yellow!20}\multicolumn{5}{l}{\textbf{(b) End-to-end latency (s)}} \\
\midrule
\multirow{2}{*}{\makecell[l]{Llama-3.1-\\70B-Instruct}}
    & LitM, $N=10$         & 57,352 & \textbf{44,219} & 22.9\% \\
    & LitM, $N=20$         & 87,091 & \textbf{58,680} & 32.6\% \\
\midrule
\multirow{3}{*}{\makecell[l]{Llama-3.1-\\8B-Instruct}}
    & MMLU, $N=4$          & 7,371  & \textbf{6,608}  & 10.4\% \\
    & LitM, $N=10$         & 18,551 & \textbf{14,264} & 23.1\% \\
    & LitM, $N=30$         & 41,664 & \textbf{23,569} & 43.4\% \\
\midrule
\rowcolor{yellow!20}\multicolumn{5}{l}{\textbf{(c) Perplexity \& Collision rate, (on LitM)}} \\
\midrule
\multirow{2}{*}{\makecell[l]{Llama-3.1-\\8B-Instruct}}
    & Perplexity ($N=20$)   & 6.91 & \textbf{6.65} & -- \\
    & Collision rate ($N=30$) & 42.3\% & \textbf{0 (None)} & -- \\
\bottomrule
\end{tabular}}
\caption{\textbf{Unified efficiency comparison of \textsc{RoToR} vs.\ PINE,} reporting (a) Additional FLOPs, (b) Latency, and (c) Perplexity \& Collision rate. Columns list each metric for PINE and RoToR, and the relative reduction.
Yellow rows separate sub-sections.}
\label{table/inference_cost}
\end{table}
Experiments against comparing \ours{} with PINE (Tab.~\ref{table/inference_cost}) we analyze \textbf{FLOPs:}~\footnote{We used the FlopsProfiler of the DeepSpeed library to measure FLOPs.} RoToR consistently reduces the floating point operations overhead across segment counts and different model backbones compared to PINE. This is because RoToR does not require computing additional attention scores: it only performs tensor operations for circular arrangement. In contrast, PINE requires more cost due to attention-based reassignment.
\textbf{Runtime, scalability:} Actual inference times (Appendix Sec.~\ref{appendix:runime_details}) find that \ours{} outperforms PINE substantially, with efficiency gains increasing alongside $n$. For instance, on LitM (30 docs), \ours{} achieves a 43\% reduction in total runtime. Practical scalability with increasing $k$ is critical, but we find that previous order-invariant LMs struggle handling larger $k$ (on KGQA and LitM). In contrast, RoToR shows better performance with improved efficiency and robustness.
\textbf{Perplexity:} Lower generation perplexity indicates input representations are closer to in-distribution. On the same LitM dataset, \ours{}’s reduced perplexity implies its positional ID assignment effectively mitigates out-of-distribution effects.
\textbf{Collision Rate:} PINE’s similarity-based ordering often collides: on average, only 17.3 of 30 similarity values are unique, causing 42\% of the segments to be indistinguishable and thus breaking invariance. In contrast, \ours{} with lexical sorting only collides if the segment texts are identical. On LitM, this yields zero collisions, preserving full invariance.

\paragraph{\sr{}} 
\label{results:other_analysis}
MMLU (Tab.~\ref{table/mmlu}) is a representative of a task that involves not only order-invariant, but also order-sensitive (e.g., "None of the above"), inputs. Therefore, single use of order-invariant models does not always outperform the original model, limiting applicability of order-invariant models to practical listwise tasks, i.e., we observe significant performance drops for Set-based Prompting in MMLU, falling short of half the performance of the original model on initial ordering. However, using \ours{} with \sr{} to handle order-sensitive inputs outperforms, or is at least competitive as the original model in all possible orderings of candidate choices. \sr{} improves the generalizability and extends the applicability on practical listwise tasks by adaptively handling order-sensitive inputs. The RoToR + \sr{} (Oracle) performance on Tab.~\ref{table/mmlu} was evaluated using a relaxed accuracy metric based on the union of predictions from the original and the invariant (RoToR-lexical) model. This improves significantly, which highlights the potential of \sr{} for further accuracy gains through optimizing design choices on routing methods, which we plan to explore in future work. Additional analysis on the selection ratio of Selective Routing is reported at Appendix Section~\ref{app:sr_stats}.

\paragraph{Impact of global ordering algorithm} While most of our experiments focus on the simplest lexical sorting method, \ours{} supports any global sorting approach. To demonstrate this flexibility, we report experiments with various global sorting strategies, including reversed lexical sorting, MonoT5-based reranking, and token frequency-based sorting. Lexical sort is presented as a baseline (lower bound) - a simple algorithm ensuring global sorting. Our experiments on Tab.~\ref{table/litm_number} show that any type of global sorting, with the use of circular assignment is \textit{superior than PINE}, which relies on pairwise attention arrangements.

\paragraph{Extension to LLMs, other scenarios}
Experiments on Llama-3.1-70B-Instruct and Qwen-1.5-72B-Chat for LitM and KGQA show consistent and significant improvements over both the original implementation and the PINE baseline, demonstrating RoToR's generalizability to larger-scale LLMs. We further evaluate robustness on longer-context inputs (LongBench-2WikiMultiHopQA~\cite{longbench} in Appendix Section~\ref{app:long_context}) and different task templates (KGQA template swap in Appendix Section~\ref{appendix:template_swap}). Results indicate that our method retains benefits across longer input scenarios and diverse task templates, including those without explicit input format requirements.

\section{Conclusion}

Our work addresses order-invariance in listwise inputs by identifying core issues in distribution mismatch and adaptive handling of mixed inputs. Our proposed RoToR by modifying self-attention by global sorting and circular arrangement provides a stable zero-shot order-invariant solution that reduces the complexity of positional ID modification, while \sr{} adaptively routes between invariant and sensitive LMs to handle real-world scenarios. Together, these methods demonstrate improved performance and reliability on LitM, KGQA, and MMLU benchmarks.

\clearpage
\newpage
\section{Limitations}
Our method can utilize any kind of deterministic sorting algorithm, but we have only experimented with limited global sorting algorithms due to time and resource constraints. We plan to investigate potentially better sorting algorithms in the future.
Also, current ordering-invariant models are limited to inputs given as prefix + parallel + suffix. It would be beneficial to support more complex structures, such as ability to process multiple order-invariant contexts interleaved with serial text.

\section{Acknowledgments}
We thank the Channel Corporation for providing GPU resources to run the experiments, and their AI team for providing helpful feedback. We thank the members of LDILab and Jinwoo Kim for their constructive comments and the anonymous reviewers for their valuable suggestions.  
We are also grateful to our former intern, Yezun Chung (KAIST CS), for assisting with experiments on early versions of the Selective Routing approach.

\noindent This work was partly supported by Institute of Information \& communications Technology Planning \& Evaluation (IITP) grant funded by the Korea government(MSIT) (No. RS-2021-II212068, Artificial Intelligence Innovation Hub) and Institute of Information \& communications Technology Planning \& Evaluation (IITP) grant funded by the Korea government(MSIT) [NO.RS-2021-II211343, Artificial Intelligence Graduate School Program (Seoul National University)]
\newpage

\bibliography{custom}

\newpage
\appendix

\onecolumn
\section*{Appendix}
\label{sec:appendix}

\section{Full results on the Lost in the Middle Benchmark}
\label{appendix:litm_detail}

\begin{figure*}[!h]
{
\centering
    
    \begin{subfigure}[b]{\textwidth}
        \centering
        \includegraphics[width=\textwidth]{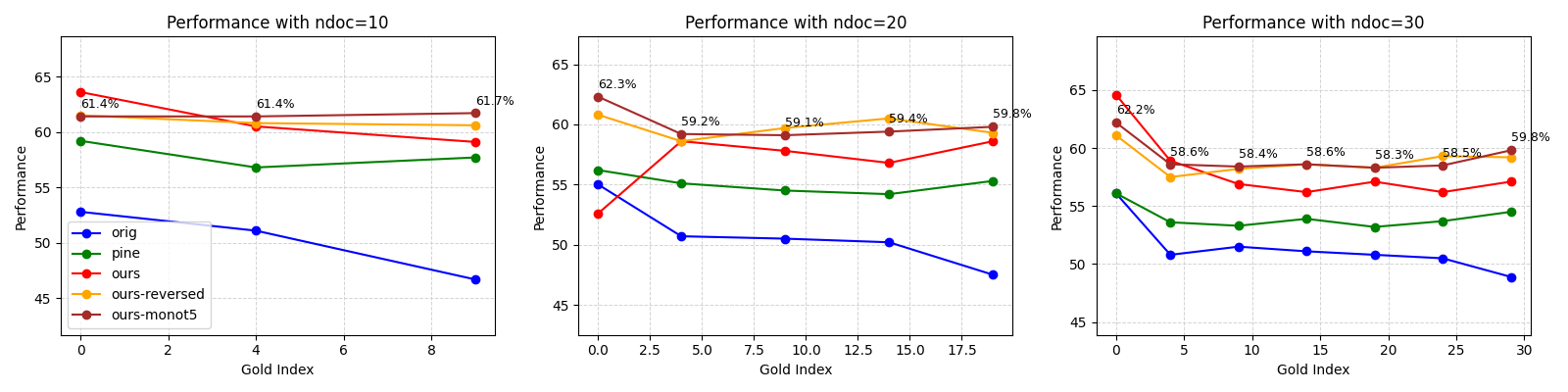}
        \caption{Results \textbf{with index bias} (indexed by numbers). (Example input at Appendix Fig.~\ref{lostinthemiddle})}
        \label{fig:litm_with_index}
    \end{subfigure}
    
    \vspace{2ex}
    
    \begin{subfigure}[b]{\textwidth}
        \centering
        \includegraphics[width=\textwidth]{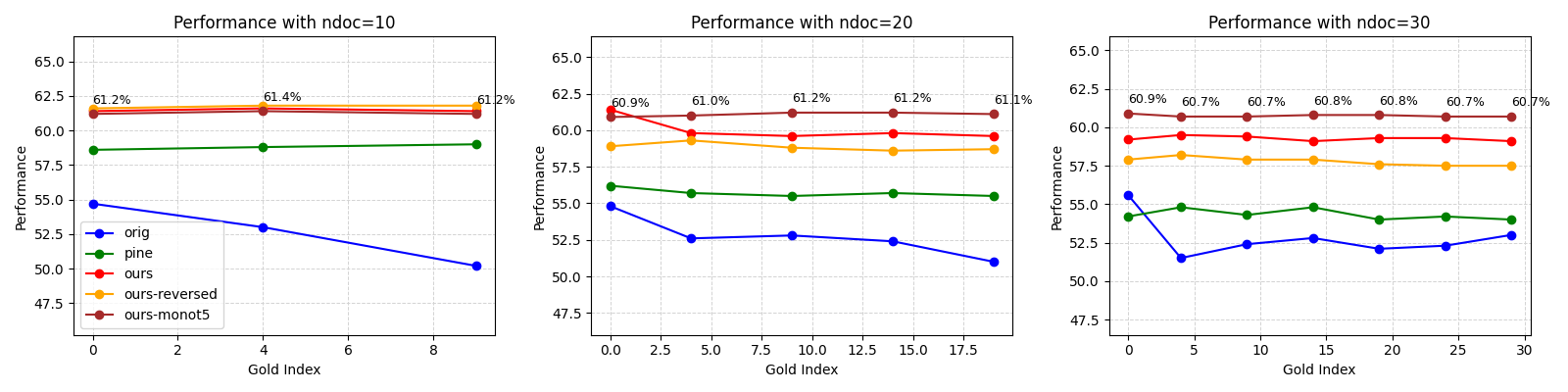}
        \caption{Results \textbf{without index bias} (indexed by title) (Example input at Appendix Fig.~\ref{lostinthemiddlenoindexing})}
        \label{fig:litm_no_index}
    \end{subfigure}
    
    \caption{Results on the Lost-in-the-middle benchmark. Visualization of the \texttt{best\_subspan\_em} results at Appendix Tab.~\ref{table/litm_number}. \ours{} (dark red, red, yellow) generally performs the best regardless of the position of the gold index, with less fluctuations when we remove index bias. Ours is \ours{} with lexical sort, and ours-reversed is the one with the reversed lexical ordering. For brevity, only the performance of \ours{} with reranking sort (MonoT5) is annotated as numbers, and the performance of PCW and Set-Based Prompting are reported only at the Table (Appendix Tab.~\ref{table/litm_number}) due to its low performance.}
    \label{fig:litm}
}
\end{figure*}
\begin{table*}[!t]

\resizebox{\linewidth}{!}
{
\begin{tabular}{@{}lcccccccccccccccccc@{}}
\toprule
\multicolumn{1}{l|}{Total ndoc (segments)} &
  \multicolumn{4}{c|}{10} &
  \multicolumn{6}{c|}{20} &
  \multicolumn{8}{c}{30} \\ \midrule
\multicolumn{1}{l|}{Gold idx at: } &
  0 &
  4 &
  9 &
  \multicolumn{1}{c|}{avg.} &
  0 &
  4 &
  9 &
  14 &
  19 &
  \multicolumn{1}{c|}{avg.} &
  0 &
  4 &
  9 &
  14 &
  19 &
  24 &
  29 &
  avg. \\ \midrule
\multicolumn{19}{l}{Indexing bias present} \\ \midrule
\multicolumn{1}{l|}{Original} &
  52.8 &
  51.1 &
  46.7 &
  \multicolumn{1}{c|}{50.2} &
  55.0 &
  50.7 &
  50.5 &
  50.2 &
  47.5 &
  \multicolumn{1}{c|}{50.7} &
  56.1 &
  50.8 &
  51.5 &
  51.1 &
  50.8 &
  50.5 &
  48.9 &
  51.4 \\
\multicolumn{1}{l|}{PCW} &
  12.0 &
  11.9 &
  12.1 &
  \multicolumn{1}{c|}{12.0} &
   \phantom{0}3.4 &
   \phantom{0}3.7 &
   \phantom{0}3.8 &
   \phantom{0}3.9 &
   \phantom{0}3.6 &
  \multicolumn{1}{c|}{ \phantom{0}5.1} &
   \phantom{0}2.1 &
   \phantom{0}2.1 &
   \phantom{0}2.0 &
   \phantom{0}1.9 &
   \phantom{0}2.0 &
   \phantom{0}2.2 &
   \phantom{0}2.0 &
   \phantom{0}2.0 \\
\multicolumn{1}{l|}{Set-Based Prompting} &
  40.8 &
  40.7 &
  40.8 &
  \multicolumn{1}{c|}{40.8} &
  25.6 &
  25.8 &
  25.7 &
  25.5 &
  25.3 &
  \multicolumn{1}{c|}{25.6} &
  15.8 &
  15.9 &
  16.1 &
  16.0 &
  16.1 &
  15.7 &
  15.8 &
  15.9 \\
\multicolumn{1}{l|}{PINE} &
  59.2 &
  56.8 &
  57.7 &
  \multicolumn{1}{c|}{57.9} &
  56.2 &
  55.1 &
  54.5 &
  54.2 &
  55.3 &
  \multicolumn{1}{c|}{55.5} &
  56.1 &
  53.6 &
  53.3 &
  53.9 &
  53.2 &
  53.7 &
  54.5 &
  54.0 \\
\multicolumn{1}{l|}{\ours{}-lexical} &
  \textbf{63.6} &
  60.5 &
  59.1 &
  \multicolumn{1}{c|}{61.1} &
  52.6 &
  58.6 &
  57.8 &
  56.8 &
  58.6 &
  \multicolumn{1}{c|}{57.6} &
  \textbf{64.6} &
  \textbf{58.9} &
  56.9 &
  56.2 &
  57.1 &
  56.2 &
  57.1 &
  58.1 \\
\multicolumn{1}{l|}{\ours{}-reversed lexical} &
  61.5 &
  60.8 &
  60.6 &
  \multicolumn{1}{c|}{61.0} &
  60.8 &
  58.6 &
  \textbf{59.7} &
  \textbf{60.5} &
  59.3 &
  \multicolumn{1}{c|}{60.0} &
  61.1 &
  57.5 &
  58.2 &
  \textbf{58.6} &
  \textbf{58.3} &
  \textbf{59.3} &
  \textbf{59.2} &
  58.9 \\ 
  \multicolumn{1}{l|}{\ours{}-reranking} &
  61.4 &
  \textbf{61.4} &
  \textbf{61.7} &
  \multicolumn{1}{c|}{\textbf{61.5}} &
  \textbf{62.3} &
  \textbf{59.2} &
   59.1 &
  59.4 &
  \textbf{59.8} &
  \multicolumn{1}{c|}{\textbf{60.2}} &
  62.2 &
  58.6 &
  \textbf{58.4} &
  \textbf{58.6} &
  \textbf{58.3} &
  58.5 &
  \textbf{59.8} &
  \textbf{59.2} \\ 
  \multicolumn{1}{l|}{\ours{}-freq} &
  62.8 &
  61.1 &
  59.5 &
  \multicolumn{1}{c|}{61.1} &
  62.9 &
  58.8 &
  56.7 &
  57.4 &
  58.0 &
  \multicolumn{1}{c|}{59.1} &
  61.7 &
  58.2 &
  56.9 &
  56.1 &
  56.4 &
  55.4 &
  56.8 &
  57.4 \\
  \midrule
\multicolumn{19}{l}{Indexing bias removed (main paper)} \\ \midrule
\multicolumn{1}{l|}{Original} &
  54.7 &
  53.0 &
  50.2 &
  \multicolumn{1}{c|}{52.6} &
  54.8 &
  52.6 &
  52.8 &
  52.4 &
  51.0 &
  \multicolumn{1}{c|}{52.7} &
  55.6 &
  51.5 &
  52.4 &
  52.8 &
  52.1 &
  52.3 &
  53.0 &
  52.8 \\
\multicolumn{1}{l|}{PCW} &
  12.4 &
  11.9 &
  12.2 &
  \multicolumn{1}{c|}{12.2} &
  \phantom{0}3.7 &
   \phantom{0}4.0 &
   \phantom{0}4.0 &
   \phantom{0}4.0 &
   \phantom{0}3.9 &
  \multicolumn{1}{c|}{ \phantom{0}3.9} &
   \phantom{0}2.3 &
   \phantom{0}1.8 &
   \phantom{0}2.0 &
   \phantom{0}2.0 &
   \phantom{0}2.1 &
   \phantom{0}2.0 &
   \phantom{0}2.0 &
   \phantom{0}2.0 \\
\multicolumn{1}{l|}{Set-Based Prompting} &
  42.5 &
  42.5 &
  42.5 &
  \multicolumn{1}{c|}{42.5} &
  26.3 &
  26.3 &
  26.3 &
  26.3 &
  26.3 &
  \multicolumn{1}{c|}{26.3} &
  14.1 &
  14.1 &
  14.1 &
  14.1 &
  14.1 &
  14.1 &
  14.1 &
  14.1 \\
\multicolumn{1}{l|}{PINE} &
  58.6 &
  58.8 &
  59.0 &
  \multicolumn{1}{c|}{58.8} &
  56.2 &
  55.7 &
  55.5 &
  55.7 &
  55.5 &
  \multicolumn{1}{c|}{55.7} &
  54.2 &
  54.8 &
  54.3 &
  53.7 &
  54.8 &
  54.2 &
  54.0 &
  54.3 \\
\multicolumn{1}{l|}{\ours{}-lexical} &
  61.4 &
  61.6 &
  61.6 &
  \multicolumn{1}{c|}{61.5} &
  \textbf{61.4} &
  59.8 &
  59.6 &
  59.6 &
  59.8 &
  \multicolumn{1}{c|}{60.0} &
59.2 &
59.5 &
59.4 &
59.1 &
59.0 &
59.3 &
59.1 &
59.2 \\
\multicolumn{1}{l|}{\ours{}-reversed lexical} &
  \textbf{61.6} &
  \textbf{61.8} &
  \textbf{61.8} &
  \multicolumn{1}{c|}{\textbf{61.8}} &
  58.9 &
  59.3 &
  58.8 &
  58.6 &
  58.7 &
  \multicolumn{1}{c|}{58.8} &
  57.9 &
  58.2 &
  57.9 &
  57.4 &
  57.9 &
  57.6 &
  57.5 &
  57.8 \\
  \multicolumn{1}{l|}{\ours{}-reranking} &
  61.2 &
  61.4 &
  61.2 &
  \multicolumn{1}{c|}{61.3} &
  60.9 &
  \textbf{61.0} &
  \textbf{61.2} &
  \textbf{61.2} &
  \textbf{61.2} &
  \multicolumn{1}{c|}{\textbf{61.1}} &
\textbf{60.9} &
\textbf{60.7} &
\textbf{60.7} &
\textbf{60.7} &
\textbf{60.8} &
\textbf{60.8} &
\textbf{60.7} &
\textbf{60.8} \\

\multicolumn{1}{l|}{\ours{}-freq} &
  61.0 &
  61.1 &
  61.1 &
  \multicolumn{1}{c|}{61.1} &
  60.4 &
  60.3 &
  58.6 &
  60.2 &
  60.0 &
  \multicolumn{1}{c|}{59.9} &
  59.3 &
  60.4 &
  59.7 &
  59.5 &
  59.5 &
  59.6 &
  59.2 &
  59.6
  \\

  \bottomrule
\end{tabular}
}

\caption{
The \texttt{best\_subspan\_em} ($\%$) scores on the lost in the middle (LitM) benchmark. For \ours{}, we test three different global ordering strategies (lexical, reversed lexical, and MonoT5-base reranking) across varying numbers of documents (ndoc $\in \{10, 20, 30\}$). Appendix Fig.~\ref{fig:litm} visualizes the fluctuations across different gold positions. \ours{} shows the best performance across all setups, and is especially more stable when indexing bias in the input is removed.
}
\label{table/litm_number_prev}

\end{table*}

\paragraph{Impact of removing index bias on LitM}
Tab.\ref{table/litm_number_prev} presents the full results on the Lost in the Middle (LitM) benchmark, comparing scenarios where indexing bias is present versus removed. Fig.\ref{fig:litm} provides a visual representation of these results.

As shown in Appendix Tab.\ref{table/litm_number_prev}, invariant LMs exhibit stable performance regardless of the gold index, especially when index bias is removed (as described in Sec.\ref{sec:benchmarks}; see also Appendix Fig.\ref{fig:litm_no_index}). However, when index bias is present, performance fluctuations are observed (Appendix Fig.\ref{fig:litm_with_index}). Notably, \ours{} achieves the highest performance across all setups, demonstrating its effectiveness in mitigating positional bias in a zero-shot setting while maintaining overall performance.

These findings suggest that index bias acts as an implicit source of additional positional bias and that invariant LMs benefit significantly from its removal.

\section{Illustration of the global sorting method}
\begin{figure}[t!]
{
\centering
    \includegraphics[width=\linewidth]{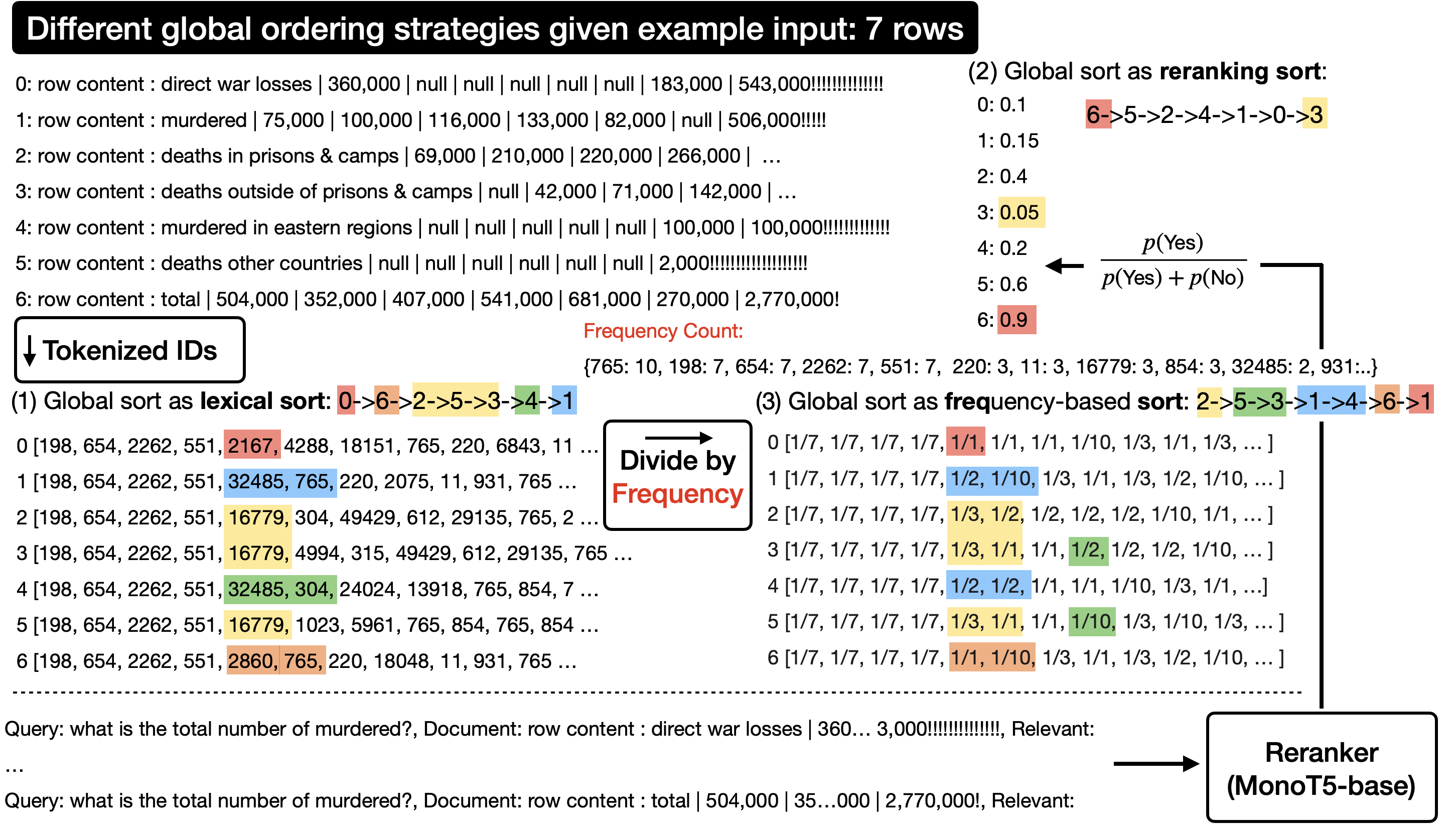}
    \caption{Illustration of ordering 7 rows by 2 different global sort options, (1) lexical sort based on token ids, or (2) reranking sort based on a reranker model (MonoT5-base in this case). (3) frequency based sorting applies lexical sorting, but the definitive token ids are mapped into the inverse frequency.}
    \label{fig:global_sort_example}
}
\end{figure}
We show the three different global sorting algorithms presented in our paper at Fig.~\ref{fig:global_sort_example}.

\section{Details about preprocessing and evaluation of datasets}
\label{appendix:dataset_details}
\subsection{General.}
\label{appendix:dataset_details_general}
All inferences were done with a \textbf{single} NVIDIA RTX A6000 48GB GPU. Note that all of the baseline models including our method can be applied directly in a zero-shot, training-free manner. For reproducibility, we fix the seed and disabled random sampling (i.e., used greedy decoding), set the maximum number of new generated tokens to 500, and set the pad\_token\_id to the same value as the eos\_token\_id for all experiments.
For all datasets we tested, we separate the input to 3 parts: prefix, parallel contexts, and suffix and feed them accordingly to the positional-invariant model. For the case of the \textbf{original} model, we simply join the prefix, context, and suffix text to make one sequential text. For testing with the \textbf{PCW} model, we concat each prefix to the parallel context due to the architectural limitations of the PCW model, which doesn't have the prefix part (which is processed casually before parallel contexts). For example, on testing the PCW model on MMLU, we append the question and each answer options, which generally result in longer input sequences. For PCW, we use the \texttt{pcw\_generate} function, and we additionally utilized the \texttt{RestrictiveTokensLogitsProcessor} provided at the official PCW repository for MMLU classification, to have a similar setup with the log\_likelihood option used for other models.

\subsection{Knowledge Graph Question Answering}
\label{appendix:dataset_details_kgqa}
For evaluation with Mintaka, 
we follow the same setup as ~\citet{baek2023knowledgeaugmentedlanguagemodelverification}. Given the gold answer and model generated answer, the EM score counts if both are exactly the same; Accuracy measures if the generated answer includes the gold answer, and F1 score measures the precision and recall among overlapping words.
Since we are testing on a non-trained zero-shot version of the model, we enforce the model to output in json format to make it easier to parse. 
For the row shuffling setup, we fix the seed to 0, 1, 2 on shuffling rows and report the average scores.

\subsection{Lost in the Middle}
\label{appendix:dataset_details_litm}
Specifically, we use the dataset provided in the official repository\footnote{\href{https://github.com/nelson-liu/lost-in-the-middle/tree/main/qa_data}{\texttt{github.com/nelson-liu/lost-in-the-middle}}}, use the same prompt as the llama 2 chat model with only the instruction tokens adjusted to llama 3 (removed \texttt{[}\texttt{Inst]} and changed to \texttt{<|begin\_of\_text|>} and etc.,), and evaluate using the \texttt{best\_supspan\_em} metric.

\subsection{MMLU}
\label{appendix:dataset_details_mmlu}
We follow the publicly acknowledged lm-evaluation harness~\cite{eval-harness} prompt design by eluther.ai. We measure accuracy between the gold answer and the token with the highest likelihood (probability) among possible answer tokens [` A’, ` B’, ` C’, ` D’].

\section{Further impact scenarios on general conversation.}
\label{appendix:discussions_about_further_impact}
We shortly discuss about how this method may be applied to general conversational scenarios of LLMs. For processing contexts such as chronological history of conversations, the ordering is important, and the original LLM remains the better choice for this case. However, in subsets of conversational tasks requiring order invariance (e.g., Sets, Tables, or RAG contexts), our method enhances unbiased understanding, as demonstrated mainly in Lost-in-the-Middle benchmark. Here, RoToR achieves a significant 7-9\% average accuracy gain over the original LLM, very consistently across all setups (doc indexing and ndoc) for all choices of the ordering algorithm, with lower standard deviation than the original model.

\section{Selection of $\alpha$ for \sr{} on MMLU}
\label{appendix:alpha_search}
\begin{table*}[!t]

\resizebox{\linewidth}{!}
{

\begin{tabular}{@{}lcccccccccccc@{}}
\toprule

\multicolumn{1}{r|}{$\alpha$ =}                         & no \sr{} & -0.5    & -0.4    & -0.3    & -0.2    & -0.1    & 0       & 0.1     & \textbf{0.2}     & 0.3    & 0.4    & 0.5    \\ \midrule
Validation set (1531)                        &        & \multicolumn{11}{c}{\textless{}--- bias towards invariant model ------ bias towards orig model ---\textgreater{}} \\ \midrule
\multicolumn{1}{l|}{\sr{} (orig, pine)}         & 64.9   & 65.3    & 65.8    & 66.4    & 67.3    & 67.4    & 67.6    & 67.5    & \textbf{67.9}    & 67.8   & 68.0   & 67.9   \\
\multicolumn{1}{l|}{\sr{} (orig, ours-lexical)} & 63.9   & 64.1    & 64.5    & 65.4    & 65.8    & 67.0    & 67.4    & 68.1    & \textbf{68.2}    & 68.1   & 67.9   & 67.9   \\
\multicolumn{1}{l|}{\sr{} (orig, ours-monot5)}  & 66.0   & 66.1    & 66.3    & 66.7    & 67.0    & 67.6    & 68.0    & 68.1    & \textbf{68.1}    & 68.1   & 67.7   & 67.9   \\ \midrule
Test set (14015)                             &        &         &         &         &         &         &         &         & \textbf{}        &        &        &        \\ \midrule
\multicolumn{1}{l|}{\sr{} (orig, pine)}         & 64.8   & 64.9    & 65.4    & 66.0    & 66.8    & 67.5    & 68.4    & 68.5    & \textbf{68.5}    & 68.4   & 68.3   & 68.3   \\
\multicolumn{1}{l|}{\sr{} (orig, ours-lexical)} & 63.2   & 63.5    & 64.2    & 65.2    & 66.2    & 67.3    & 68.0    & 68.4    & \textbf{68.5}    & 68.5   & 68.4   & 68.3   \\
\multicolumn{1}{l|}{\sr{} (orig, ours-monot5)}  & 64.2   & 64.4    & 64.8    & 65.5    & 66.4    & 67.3    & 68.1    & 68.5    & \textbf{68.4}    & 68.5   & 68.3   & 68.3   \\ \bottomrule
\end{tabular}

}

\caption{Reporting full ablation results on application of Selective Routing. $\alpha$ = 0.2 was the best for the validation set, which was then applied to obtain the reported results for all models.}
\label{table/mov_alpha_ablation}

\end{table*}

We report 
$\alpha$ is a hyperparameter that can be tuned per-dataset. We searched its value in the range of -0.5 to 0.5 with a step size of 0.1 using the validation split of MMLU\footnote{\url{https://huggingface.co/datasets/cais/mmlu/viewer/abstract_algebra/validation}} on RoToR with lexical sorting, and applied the found value (0.2) on the test set to obtain the reported results for all models. We report the full variation of \sr{} results on the investigated $\alpha$ value at Tab.~\ref{table/mov_alpha_ablation}.

\section{Replication details on the runtime experiment}
\label{appendix:runime_details}
Apart from the theoretical runtime efficiency, we measured the actual end-to-end runtime in seconds, to better analyze the practical runtime efficiency between PINE and \ours{}.
The runtime of each experiment was measured on an ASUS ESC8000-E11 server featuring dual 4th Gen Intel Xeon Scalable processors, 64 CPU threads, 1.1 TB of RAM across 32 DIMM slots, and 8 NVIDIA A6000 GPUs with 48 GB of memory each. We  Except for the experiments on Llama-3.1-70B-Instruct, we only use a single A6000 GPU for all of the experiments.

\section{Input data examples}
\label{appendix:inputexamples}
To illustrate the input and output formats used in our experiments, we provide example inputs for the Lost-in-the-Middle (LitM), Knowledge Graph Question Answering (KGQA), and MMLU datasets. For experiments using the Qwen-Chat model, special tokens were adjusted accordingly. While the example prompts are based on the Llama-3.1-8B-Instruct model, the specific differences in token usage for the Qwen-Chat variants can be observed by comparing the prompts in Fig.~\ref{lostinthemiddlenoindexing} and Fig.~\ref{lostinthemiddlenoindexing_qwen}. This adjustment is consistently applied across all datasets. Note that no special tokens are added for the MMLU benchmark, which aligns with the \texttt{lm-evaluation harness} setup.

\begin{tcolorbox}[title=lost in the middle]
\vspace{-0.2cm}
\textbf{Prefix:}\\
<|begin\_of\_text|><|start\_header\_id|>system<|end\_header\_id|>\\

You are a helpful, respectful and honest assistant. Always answer as helpfully as possible, while being safe. Please ensure that your responses are socially unbiased and positive in nature. If a question does not make any sense, or is not factually coherent, explain why instead of answering something not correct. If you don't know the answer to a question, please don't share false information.<|eot\_id|><|start\_header\_id|>user<|end\_header\_id|>\\

Write a high-quality answer for the given question using only the provided search results (some of which might be irrelevant).\\

\textbf{Parallel texts:}\\
Document [1](Title: List of Nobel laureates in Physics) The first ... \\
...\\
Document [10](Title: Nobel Prize in Chemistry) on December 10, the ... \\

\textbf{Suffix: }\\
Question: who got the first nobel prize in physics<|eot\_id|><|start\_header\_id|>assistant\\<|end\_header\_id|>
\end{tcolorbox}
\noindent\begin{minipage}{\textwidth}
\captionof{figure}{Example input for the lost in the middle dataset.}\label{lostinthemiddle}
\end{minipage}

\newpage

\begin{tcolorbox}[title=lost in the middle no indexing]
\textbf{Prefix:}\\
<|begin\_of\_text|><|start\_header\_id|>system<|end\_header\_id|>\\

You are a helpful, respectful and honest assistant. Always answer as helpfully as possible, while being safe. Please ensure that your responses are socially unbiased and positive in nature. If a question does not make any sense, or is not factually coherent, explain why instead of answering something not correct. If you don't know the answer to a question, please don't share false information.<|eot\_id|><|start\_header\_id|>user<|end\_header\_id|>\\

Write a high-quality answer for the given question using only the provided search results (some of which might be irrelevant).\\

\textbf{Parallel texts:}\\
\texttt{[}Document Title: List of Nobel laureates in Physics\texttt{]} The first ... \\
...\\
\texttt{[}Document Title: Nobel Prize in Chemistry\texttt{]} on December 10, the ... \\

\textbf{Suffix: }\\
Question: who got the first nobel prize in physics<|eot\_id|><|start\_header\_id|>assistant\\<|end\_header\_id|>
\end{tcolorbox}

\noindent\begin{minipage}{\textwidth}
\captionof{figure}{Example input for the lost in the middle dataset, without indexing by numbers. Prompt for the Llama-3.1-8B-Instruct model.}\label{lostinthemiddlenoindexing}
\end{minipage}

\begin{tcolorbox}[title=lost in the middle no indexing (Qwen variant)]
\textbf{Prefix:}\\
<|im\_start|>system\\
You are a helpful, respectful and honest assistant. Always answer as helpfully as possible, while being safe. Please ensure that your responses are socially unbiased and positive in nature. If a question does not make any sense, or is not factually coherent, explain why instead of answering something not correct. If you don't know the answer to a question, please don't share false information.<|im\_end|><|im\_start|>user\\

Write a high-quality answer for the given question using only the provided search results (some of which might be irrelevant).\\

\textbf{Parallel texts:}\\
\texttt{[}Document Title: Thorax\texttt{]} when deep breaths are attempted. Different people ...
\\
...\\
\texttt{[}Document Title: Chest pain\texttt{]} present with chest pain, and carry a significantly higher  ... \\

\textbf{Suffix: }\\
Question: for complaints of sudden chest pain patients should take a<|im\_end|>\\
<|im\_start|>assistant\\

\end{tcolorbox}

\noindent\begin{minipage}{\textwidth}
\captionof{figure}{Example input for the lost in the middle dataset, without indexing by numbers, prompt for the Qwen1.5-Chat model.}\label{lostinthemiddlenoindexing_qwen}
\end{minipage}

\newpage

\begin{tcolorbox}[title=Mintaka]
\textbf{Prefix:}\\
<|begin\_of\_text|><|start\_header\_id|>system<|end\_header\_id|>\\

Below are the facts in the form of the triple meaningful to answer the question. Answer the given question in a JSON format, such as {"Answer": "xxx"}. Only output the JSON, do NOT say any word or explain.\\

<|eot\_id|><|start\_header\_id|>user<|end\_header\_id|>\\

\textbf{Parallel texts:}\\
(Super Bowl XLII, winner, New York Giants) \\
(Super Bowl XLII, participating team, New York Giants) \\
(Super Bowl XLII, point in time, time: +2008-02-03) \\
(Super Bowl XLII, followed by, Super Bowl XLIII) \\
(Super Bowl XLII, location, State Farm Stadium) \\
... \\
(Super Bowl XLII, sport, American football) \\
(Super Bowl XLII, instance of, Super Bowl) \\

\textbf{Suffix: }\\
Question: which team did the super bowl xlii mvp play for?, Answer: <|eot\_id|><|start\_header\_id|> assistant 
<|end\_header\_id|> \\

\textbf{Gold Answer(s): }\\
(`NYG', `Giants', `NY Giants', `New York Giants') \\

\textbf{Example generated output: }\\
\{"Answer": "New York Giants"\} (Parsed to: New York Giants)

\end{tcolorbox}

\noindent\begin{minipage}{\textwidth}
\captionof{figure}{Example input for the Mintaka dataset.}\label{Mintaka}
\end{minipage}

\begin{tcolorbox}[title=MMLU]
\vspace{-0.2cm}
\textbf{Prefix:}\\
The following are multiple choice questions (with answers) about moral disputes.\\

Norcross agrees that if a being is incapable of moral reasoning, at even the most basic level, then it cannot be\\

\textbf{Parallel texts:}\\
A. a being of value. \\
B. an object of moral sympathy. \\
C. a moral agent. \\
D. a moral patient. \\

\textbf{Suffix: }\\
Answer:
\end{tcolorbox}
\noindent\begin{minipage}{\textwidth}
\captionof{figure}{Example input for the MMLU benchmark.}\label{mmlu_prompt}
\end{minipage}

\newpage

\section{Why is removing index bias an important detail for invariant models to be effective?}
\label{appendix:index_bias}

The alphabetic index (A/B/C/D) introduced in Fig.~\ref{fig:intro_example} associated with each segment, reportedly introduces token bias~\cite{wei2024unveilingselectionbiasesexploring} of preferring the choice marked as `A.' The same thing can be applied to listwise inputs with simple numeric indexing (1/2/3/4), which was the case for the lost-in-the middle benchmark. While a standard model with no modifications on positional encoding correctly places contexts indexed A before contexts indexed with D by positional encoding, an invariant model sees contexts in an order-agnostic way, meaning that the alphabetical indexing may not always be interpreted sequentially and thus can confuse the model from accurately interpreting the contexts.
For example, even for cases where the index ordering of the input was in alphabetical order (A->B->C->D), the ordering-invariant model may interpret contexts with (C->A->B->D) at one point (e.g., when the query is D on self attention), which can cause unnatural, out-of-distribution representation, leading to decreased performance.

\section{Statisticial significance before and after shuffling segments}
\label{appendix:statistical_significance}

We conducted \textbf{paired two-tailed \textit{t}-tests} (Table~\ref{tab:t_test_summary}) for both the baseline (``original'') model and our proposed method (\ours{}), using the results in Table~\ref{tab:t_test}. Our goal was to determine whether the performance differences between the initial ordering and shuffled ordering are statistically significant. We excluded the Lost-in-the-Middle (LitM) dataset because it does not provide an initial ordering. Specifically, the tests evaluate whether the mean performance difference (\text{Before Shuffle} - \text{After Shuffle}) significantly deviates from zero.

For KGQA, we selected the F1 score as the representative metric among the three available, gathering data points from various task configurations and different models. For MMLU, the results are based on our \ours{} variant with selective routing. As shown in Table~\ref{tab:t_test_summary}, the original model shows a statistically significant drop in performance when the segments are shuffled, while \ours{} does not, indicating increased robustness to segment-order perturbations. \footnote{All statistical calculations were validated using an online t-test calculator: \url{https://www.mathportal.org/calculators/statistics-calculator/t-test-calculator.php}}

\begin{table}[h!]
\centering

\resizebox{0.95\linewidth}{!}
{
\begin{tabular}{@{}l|ccc|ccc@{}}
\toprule
 & \multicolumn{3}{c|}{Original Model} & \multicolumn{3}{c}{RoToR-lexical} \\ \midrule
 & \begin{tabular}[c]{@{}c@{}}Before \\ Shuff.\end{tabular} & \begin{tabular}[c]{@{}c@{}}After \\ Shuff.\end{tabular} & Diff. & \begin{tabular}[c]{@{}c@{}}Before \\ Shuff.\end{tabular} & \begin{tabular}[c]{@{}c@{}}After \\ Shuff.\end{tabular} & Diff. \\ \midrule
Mintaka, Llama3.1-8B-Instruct, ndoc=30 & 51.9 & 51.0 & 0.9 & 54.1 & 53.8 & 0.3 \\
Mintaka, Llama3.1-8B-Instruct, ndoc=50 & 51.7 & 51.0 & 0.7 & 53.6 & 53.5 & 0.1 \\
Mintaka, Qwen1.5-4B-Chat, ndoc=30 & 34.9 & 34.7 & 0.2 & 35.7 & 35.5 & 0.2 \\
Mintaka, Qwen1.5-4B-Chat, ndoc=50 & 35.8 & 35.0 & 0.8 & 36.2 & 36.1 & 0.1 \\
Mintaka, Qwen1.5-7B-Chat, ndoc=30 & 35.4 & 35.0 & 0.4 & 37.7 & 37.7 & 0 \\
Mintaka, Qwen1.5-7B-Chat, ndoc=50 & 35.7 & 35.5 & 0.2 & 38.0 & 38.0 & 0 \\
MMLU, Llama3.1-8B-Instruct & 68.3 & 65.5 & 2.8 & 68.5 & 65.7 & 2.8 \\
MMLU, Qwen1.5-4B-Chat & 53.6 & 52.6 & 1 & 53.7 & 52.6 & 1.1 \\
MMLU, Qwen1.5-7B-Chat & 60.1 & 58.6 & 1.5 & 60.1 & 58.8 & 1.3 \\ \bottomrule
\end{tabular}
}
\caption{Performance of the \textbf{Original model} and \textbf{\ours{}} before and after shuffling.}
\label{tab:t_test}
\end{table}

\begin{table}[ht]
\centering
\begin{tabular}{@{}l|cc@{}}
\toprule
 & \multicolumn{1}{c|}{\textbf{Original}} & \textbf{RoToR} \\ \midrule
Degrees of Freedom & \multicolumn{2}{c}{8} \\ \midrule
Mean Difference & \multicolumn{1}{c|}{0.94} & 0.66 \\ \midrule
\textit{t}-Statistic & \multicolumn{1}{c|}{3.71} & 2.23 \\ \midrule
Critical Value & \multicolumn{2}{c}{2.306} \\ \midrule
Statistically Significant & \multicolumn{1}{c|}{\textbf{Yes}} & \textbf{No} \\ \bottomrule
\end{tabular}
\caption{Paired two-tailed t-test results comparing the original model and ours.}
\label{tab:t_test_summary}
\end{table}

\paragraph{Derivation for the original model.}
Let the nine paired differences (Before~$-$ After) be \(\{d_1, d_2, d_3, ... d_8, d_9\}\).  
\textbf{Mean Difference:} $\bar{d} = \tfrac{1}{9}\sum_{i=1}^{9} d_i$. In this case, $\bar{d} \approx 0.9444\%.$  
\textbf{Sample Standard Deviation:} $s_d = \sqrt{\frac{1}{n-1}\sum_{i=1}^{n} (d_i - \bar{d})^2} \approx 0.7632.$  
\textbf{Standard Error (SE):} $\mathrm{SE} = \tfrac{s_d}{\sqrt{n}} \approx 0.2544.$  
\textbf{\textit{t}-Statistic:} $t = \tfrac{\bar{d}}{\mathrm{SE}} \approx 3.7124,\ (df=8).$  
Since the critical value at $df=8$ and $\alpha=0.05$ is $2.306,$ we have $3.71 > 2.306.$ Therefore, the difference is statistically significant.

\paragraph{Derivation for the \ours{} model.}
Under the same procedure, $\bar{d} \approx 0.6556\%,\ s_d \approx 0.8833,\ \mathrm{SE} \approx 0.2944.$  
\textbf{\textit{t}-Statistic:} $t \approx 2.2265.$ Since $2.2265 < 2.306,$ there is no significant difference in performance before and after shuffling for \ours{}.

\paragraph{Conclusion.}
While the original model shows a statistically significant performance drop with shuffled inputs, \ours{} remains unaffected, demonstrating greater robustness to segment-order perturbations.

\section{Application to Long-Context Inputs}
\label{app:long_context}

\paragraph{Benchmark and protocol.} To test whether RoToR consistantly performs well to inputs with longer contexts, we extend our evaluation to the \textbf{LongBench}~\cite{longbench}–\textbf{2WikiMultihopQA} task, whose
multi-document questions yield input lengths from 5 k to 15 k tokens.
Because our current \textsc{RoToR} and \textsc{PINE} implementations do \emph{not} yet
support advanced parallelization (e.g., FlashAttention or SDPA),
GPU memory becomes prohibitive beyond $\sim$10 k tokens, especially for
\textsc{PINE}, whose memory footprint is larger than ours.
We therefore truncate the context at~10 k tokens, which already exceeds
the lengths used in our main experiments.

Following the official LongBench guidelines\footnote{%
Maximum generation length~32 and
\texttt{qa\_f1\_score} (LongBench-E) as the evaluation
metric.},
we segment each long context into 512-token “listwise” inputs,
enabling a direct comparison with our listwise reranking pipeline.
To ensure robustness, we test three input-order perturbations:

\begin{enumerate}
    \item \textbf{Initial}: original chunk order \([1,2,3,4,5]\);
    \item \textbf{Reversed}: \([5,4,3,2,1]\);
    \item \textbf{Center-flipped}: first and last halves swapped,\,[3,2,1,5,4].
\end{enumerate}

\vspace{-0.5em}
\paragraph{Results.}
Table~\ref{tab:longbench_2wiki} reports F1~scores for
the Original model variant, RoToR, and PINE 
with \textbf{Llama 3.1-8B-Instruct} and
\textbf{Qwen 1.5-7B-Chat}.
Across every ordering and token-length band, RoToR retains a clear advantage,
while PINE fails to run when the longest (8 k+) chunks are kept.
These findings confirm that RoToR scales to substantially
longer inputs across different input-order perturbations.

\begin{table*}[t]
\centering
\small
\setlength{\tabcolsep}{4pt}   
\begin{tabular}{ll|cccc|cccc}
\toprule
      &       & \multicolumn{4}{c|}{\textbf{Llama 3.1-8B-Instruct}} & \multicolumn{4}{c}{\textbf{Qwen 1.5-7B-Chat}} \\
\cmidrule(lr){3-6}\cmidrule(lr){7-10}
Order & Method & 0--4k & 4--8k & 8k+ & Total & 0--4k & 4--8k & 8k+ & Total \\ \midrule
      & Count  & 25 & 131 & 144 & 300 & 23 & 121 & 156 & 300 \\ \midrule
\multirow{3}{*}{\shortstack[l]{Initial\\(e.g., \\ 1,2,3,4,5)}} 
  & Orig.  & 48.3 & 56.8 & 34.0 & 45.1 & 65.6 & 47.9 & 26.7 & 38.2 \\
  & PINE   & 51.0 & 47.6 & --   & --   & 70.2 & 45.1 & --   & --   \\
  & RoToR  & \textbf{59.0} & 52.7 & \textbf{41.8} & \textbf{48.0}
           & \textbf{75.7} & \textbf{47.8} & \textbf{31.0} & \textbf{41.2} \\ \midrule
\multirow{3}{*}{\shortstack[l]{Reversed\\(e.g., \\5,4,3,2,1)}} 
  & Orig.  & 57.0 & 51.5 & 39.0 & 46.0 & 53.4 & 43.3 & \textbf{34.2} & 39.3 \\
  & PINE   & 43.0 & 49.8 & --   & --   & 64.1 & \textbf{48.9} & --   & --   \\
  & RoToR  & \textbf{59.0} & \textbf{52.0} & \textbf{41.0} & \textbf{47.3}
           & \textbf{72.8} & 47.6 & 30.8 & \textbf{40.8} \\ \midrule
\multirow{3}{*}{\shortstack[l]{Center flip\\(e.g., \\3,2,1,5,4)}} 
  & Orig.  & 47.0 & 47.7 & 35.6 & 41.8 & 61.0 & 40.6 & 32.7 & 38.1 \\
  & PINE   & 46.3 & 49.2 & --   & --   & 70.2 & 43.5 & --   & --   \\
  & RoToR  & \textbf{59.0} & \textbf{52.5} & \textbf{41.5} & \textbf{47.8}
           & \textbf{77.1} & \textbf{47.3} & \textbf{30.9} & \textbf{41.0} \\ 
\bottomrule
\end{tabular}
\caption{F1 scores (\%) on \textsc{LongBench}--2WikiMultihopQA with $\sim$10k-token contexts.
``Count'' is the number of examples per length bucket; ``--'' denotes out-of-memory.}
\label{tab:longbench_2wiki}
\end{table*}

\paragraph{Take-aways.}
Even without specialised long-context kernels,
\textsc{RoToR} consistently outperforms both
the original reranker and \textsc{PINE},
and remains robust to severe input-order perturbations.
This suggests our approach can generalize to
substantially longer inputs once memory
and kernel constraints are alleviated.

\section{Robustness to task templates}
\label{appendix:template_swap}

\begin{table*}[t]
\centering
\resizebox{0.95\linewidth}{!}{
\begin{tabular}{@{}lcccccccccccccccccc@{}}
\toprule
\multicolumn{1}{l|}{} & \multicolumn{6}{c|}{Llama-3.1-8B-Instruct} & \multicolumn{6}{c|}{Qwen-1.5-4B-Chat} & \multicolumn{6}{c}{Qwen-1.5-7B-Chat} \\ \midrule
\multicolumn{1}{l|}{} & \multicolumn{3}{c|}{$N=30$} & \multicolumn{3}{c|}{$N=50$} & \multicolumn{3}{c|}{$N=30$} & \multicolumn{3}{c|}{$N=50$} & \multicolumn{3}{c|}{$N=30$} & \multicolumn{3}{c}{$N=50$} \\ \cmidrule(l){2-19}
\multicolumn{1}{l|}{\textbf{Method}} & Acc. & EM & \multicolumn{1}{c|}{F1} & Acc. & EM & \multicolumn{1}{c|}{F1} & Acc. & EM & \multicolumn{1}{c|}{F1} & Acc. & EM & \multicolumn{1}{c|}{F1} & Acc. & EM & \multicolumn{1}{c|}{F1} & Acc. & EM & F1 \\ \midrule
\multicolumn{19}{l}{\textbf{Original template, Initial ordering}} \\ \midrule
\multicolumn{1}{l|}{Original}         & 50.2 & 44.0 & \multicolumn{1}{c|}{51.9} & 50.0 & 44.0 & \multicolumn{1}{c|}{51.7} & 30.7 & 27.9 & \multicolumn{1}{c|}{34.9} & 31.6 & 28.6 & \multicolumn{1}{c|}{35.8} & 31.5 & 27.8 & \multicolumn{1}{c|}{35.4} & 31.7 & 28.0 & 35.7 \\
\multicolumn{1}{l|}{PINE}             & 51.5 & 45.0 & \multicolumn{1}{c|}{52.6} & 51.6 & 45.1 & \multicolumn{1}{c|}{52.6} & 31.6 & 28.7 & \multicolumn{1}{c|}{35.6} & 31.6 & 28.8 & \multicolumn{1}{c|}{35.3} & 32.3 & 28.8 & \multicolumn{1}{c|}{36.4} & 32.0 & 28.5 & 35.9 \\
\multicolumn{1}{l|}{RoToR}            & \textbf{53.1} & \textbf{46.5} & \multicolumn{1}{c|}{\textbf{54.1}} & 52.9 & 46.0 & \multicolumn{1}{c|}{53.6} & 32.0 & 29.0 & \multicolumn{1}{c|}{35.7} & \textbf{32.7} & \textbf{29.6} & \multicolumn{1}{c|}{\textbf{36.2}} & \textbf{34.3} & \textbf{29.8} & \multicolumn{1}{c|}{\textbf{37.7}} & \textbf{34.3} & \textbf{30.1} & \textbf{38.0} \\
\multicolumn{1}{l|}{RoToR-MonoT5}     & 51.6 & 45.0 & \multicolumn{1}{c|}{52.5} & 52.4 & 45.4 & \multicolumn{1}{c|}{52.8} & \textbf{32.3} & 29.1 & \multicolumn{1}{c|}{\textbf{36.2}} & 32.3 & 29.3 & \multicolumn{1}{c|}{35.9} & 32.9 & 28.4 & \multicolumn{1}{c|}{36.3} & 32.9 & 28.9 & 36.6 \\
\multicolumn{1}{l|}{RoToR-Freq.}      & 52.6 & 46.1 & \multicolumn{1}{c|}{53.7} & \textbf{53.1} & \textbf{46.4} & \multicolumn{1}{c|}{\textbf{53.7}} & \textbf{32.3} & \textbf{29.2} & \multicolumn{1}{c|}{36.0} & 32.3 & 29.2 & \multicolumn{1}{c|}{35.9} & 33.7 & 29.5 & \multicolumn{1}{c|}{37.2} & 33.5 & 29.5 & 37.2 \\ \midrule
\multicolumn{19}{l}{\textbf{Template-swap}} \\ \midrule
\multicolumn{1}{l|}{Original}         & 50.0 & 44.1 & \multicolumn{1}{c|}{51.8} & 50.2 & 44.3 & \multicolumn{1}{c|}{51.9} & 31.1 & 27.8 & \multicolumn{1}{c|}{34.9} & 31.7 & 28.3 & \multicolumn{1}{c|}{35.3} & 31.4 & 27.6 & \multicolumn{1}{c|}{35.2} & 32.0 & 28.0 & 35.7 \\
\multicolumn{1}{l|}{PINE}             & 52.0 & 45.7 & \multicolumn{1}{c|}{52.9} & 52.0 & 45.3 & \multicolumn{1}{c|}{52.8} & 31.9 & 28.8 & \multicolumn{1}{c|}{35.8} & 31.7 & 28.6 & \multicolumn{1}{c|}{35.4} & 31.9 & 28.5 & \multicolumn{1}{c|}{36.0} & 31.5 & 28.2 & 35.7 \\
\multicolumn{1}{l|}{RoToR}            & \textbf{52.7} & \textbf{46.4} & \multicolumn{1}{c|}{\textbf{54.0}} & \textbf{52.9} & 46.4 & \multicolumn{1}{c|}{53.7} & 31.8 & 28.2 & \multicolumn{1}{c|}{35.0} & 32.4 & 29.0 & \multicolumn{1}{c|}{35.6} & \textbf{34.1} & 29.8 & \multicolumn{1}{c|}{\textbf{37.6}} & \textbf{34.0} & \textbf{29.9} & \textbf{37.7} \\
\multicolumn{1}{l|}{RoToR-MonoT5}     & 51.5 & 45.2 & \multicolumn{1}{c|}{52.6} & 52.5 & 45.7 & \multicolumn{1}{c|}{53.1} & \textbf{32.4} & \textbf{29.0} & \multicolumn{1}{c|}{\textbf{36.3}} & \textbf{32.6} & \textbf{29.3} & \multicolumn{1}{c|}{\textbf{35.9}} & 32.9 & 28.5 & \multicolumn{1}{c|}{36.4} & 32.6 & 28.5 & 36.3 \\
\multicolumn{1}{l|}{RoToR-Freq.}      & 52.3 & 46.2 & \multicolumn{1}{c|}{53.7} & \textbf{52.9} & \textbf{46.5} & \multicolumn{1}{c|}{\textbf{53.8}} & 31.9 & 28.4 & \multicolumn{1}{c|}{35.4} & 32.4 & 28.8 & \multicolumn{1}{c|}{35.6} & 34.0 & \textbf{29.9} & \multicolumn{1}{c|}{37.4} & 33.8 & 29.6 & 37.4 \\ \bottomrule
\end{tabular}}
\caption{Results on the Mintaka (KGQA) dataset on different models, before (top block, also reported at main paper) and after (bottom block) the template-swap.  N refers to number of top-k segments per query. RoToR variants consistently outperform the Original and PINE baselines, and their performance is stable under the swapped template, indicating robustness to instruction wording.}
\label{tab:template_swap}
\end{table*}
We evaluated whether the proposed selective-routing methods remain effective when the surrounding task instructions are re-phrased. Concretely, we performed a \emph{template-swap} experiment on the KGQA benchmark, specifically on the initial ordering setup.
The original prompt (Figure~\ref{Mintaka}) began with  
\begin{quote}
\small
“Below are the facts in the form of triples meaningful to answer the question.”
\end{quote}
and required the model to output only a JSON object.  
In the swapped template we replaced the first sentence with  
\begin{quote}
\small
“Below are knowledge statements expressed as triples meaningful to answer the question.”
\end{quote}
leaving all other instructions unchanged.  
Table \ref{tab:template_swap} reports the results.  Across all three backbone models and both retrieval depths ($N{=}30,50$), RoToR and its variants retain similar absolute scores and continue to outperform both the Original and PINE baselines, indicating strong robustness to superficial wording changes in the task template.

\section{Additional Statistics on Selective Routing Assignment}
\label{app:sr_stats}

Table~\ref{tab:sr_stats} complements the main results in Table~\ref{table/mmlu} by reporting the \emph{selection ratio}, accounting for the percentage of evaluation queries for which the RoToR branch is chosen over the vanilla branch, under \textbf{Selective Routing} (SR).\footnote{All figures are computed over the full evaluation set of 14,015 queries.}  We break the analysis down by (i) the global sorting strategy (Lexical, MonoT5, or Freq.), and (ii) three model backbones.  The table distinguishes three order–based conditions:

\begin{table}[h]
\centering
\small
\setlength{\tabcolsep}{4pt}
\begin{tabular}{lccc ccc ccc}
\toprule
 & \multicolumn{3}{c}{\textbf{Llama-3.1-8B-Instr.}} & \multicolumn{3}{c}{\textbf{Qwen1.5-4B-Chat}} & \multicolumn{3}{c}{\textbf{Qwen1.5-7B-Chat}}\\
\cmidrule(lr){2-4}\cmidrule(lr){5-7}\cmidrule(lr){8-10}
\textbf{Sorting} & Init. & Rev. & Avg. & Init. & Rev. & Avg. & Init. & Rev. & Avg.\\
\midrule
Lexical & 7.0 & 8.5 & $7.3\!\pm\!0.8$ & 5.9 & 6.2 & $6.2\!\pm\!0.4$ & 10.3 & 10.6 & $9.9\!\pm\!0.6$ \\
MonoT5  & 6.9 & 7.6 & $6.7\!\pm\!1.5$ & 8.0 & 12.5 & $9.8\!\pm\!2.1$ & 10.7 & 10.9 & $10.7\!\pm\!0.7$ \\
Freq.   & 6.4 & 6.7 & $6.9\!\pm\!0.5$ & 8.5 & 10.9 & $9.4\!\pm\!1.6$ & 10.7 & 11.1 & $11.1\!\pm\!0.8$ \\
\bottomrule
\end{tabular}
\caption{Selection ratio (\%) of the RoToR variant under SR.  Higher values indicate more frequent routing to RoToR.}
\label{tab:sr_stats}
\end{table}

For table~\ref{tab:sr_stats}, Init. refers to the original ordering (e.g., in abcd order), Rev. refers to the reversed ordering (e.g., in dcba but assigned as abcd), and Avg. is the average selection ratio for all possible (4!-1) re-orderings, with standard deviation.
Empirically, the RoToR model tends to be selected more frequently under reversed orderings, whereas under the original ordering, the vanilla model is chosen slightly more often. The exact ratio varies by model and sorting strategy.

\end{document}